\documentclass{article}

\usepackage{arxiv}

\usepackage[T1]{fontenc}    
\usepackage{booktabs}       
\usepackage{nicefrac}       
\usepackage{microtype}      
\usepackage{graphicx}
\usepackage{natbib}
\usepackage{doi}

\usepackage[utf8]{inputenc}
\usepackage{url}
\usepackage{hyperref}
\usepackage{amsmath,amsfonts,amssymb}
\usepackage[english]{babel}
\usepackage{amsthm}
\usepackage{cleveref}
\usepackage{caption}
\usepackage{subcaption}
\usepackage{amsmath}
\usepackage{xcolor}
\usepackage{stackengine}
\usepackage{mathtools}
\usepackage{placeins}
\usepackage{multirow}
\usepackage{wrapfig}
\usepackage{lipsum}

\title{Linear Log-Normal Attention with Unbiased Concentration}

\date{} 					

\author{Yury Nahshan, Joseph Kampeas and Emir Haleva \\
            Distributed and Parallel Software Lab, Huawei Technologies \\
            \texttt{Email: \{first.last\}@huawei.com}
            }

\hypersetup{
pdftitle={Linear Log-Normal Attention with Unbiased Concentration},
pdfsubject={paper},
pdfauthor={Yury Nahshan, Joseph Kampeas and Emir Haleva},
pdfkeywords={First keyword, Second keyword, More},
}

\begin{document}
\maketitle

\begin{abstract}
Transformer models have achieved remarkable results in a wide range of applications. However, their scalability is hampered by the quadratic time and memory complexity of the self-attention mechanism concerning the sequence length. This limitation poses a substantial obstacle when dealing with long documents or high-resolution images. In this work, we study the self-attention mechanism by analyzing the distribution of the attention matrix and its concentration ability. Furthermore, we propose instruments to measure these quantities and introduce a novel self-attention mechanism, Linear Log-Normal Attention, designed to emulate the distribution and concentration behavior of the original self-attention. Our experimental results on popular natural language benchmarks reveal that our proposed Linear Log-Normal Attention outperforms other linearized attention alternatives, offering a promising avenue for enhancing the scalability of transformer models.
\end{abstract}

\newcommand\BB[1]{\textcolor{blue}{#1}} 
\newcommand\OB[1]{\textcolor{orange}{#1}}

\newcommand{\innerproduct}[2]{\left\langle #1, #2 \right\rangle}
\newcommand{\tran}[1]{{#1 ^ \intercal}}
\newcommand{\mb}[1]{{\pmb{#1}}}
\newcommand{\con}[1]{{\overline{#1}}}
\newcommand{\norm}[1]{{\left\|{#1}\right\|}}
\newcommand{\bls}[1]{{\boldsymbol{#1}}}
\newcommand{\polykernel}[3]{\innerproduct{#1}{#2}^#3}
\newcommand{\derv}[2]{\frac{\partial #1}{\partial #2}}

\newtheorem{theorem}{Theorem}[section]
\newtheorem{corollary}{Corollary}[theorem]
\newtheorem{lemma}[theorem]{Lemma}
\newtheorem{proposition}[theorem]{Proposition}
\newtheorem{definition}[theorem]{Definition}
\newtheorem{claim}[theorem]{Claim}

\newtheorem*{model_def}{Model Definition}

\section{Introduction}
Transformer models, proposed by \citep{VaswaniSPUJGKP17}, have become widely used deep learning architectures that have achieved state-of-the-art performance in various fields, including Natural Language Processing (NLP) \citep{brown2020language, Bert-2018}, Computer Vision (CV) \citep{Vision-16-16}, Neural Machine Translation (NMT) \citep{NMT_best_of_words}, Document Summarization \citep{hilbert-doc-sum, pilault-etal-2020-extractive}, and Protein Structure Prediction \citep{attention-bahdanau}. The main component of the Transformer model is an attention mechanism that identifies complex dependencies between tokens and efficiently captures tokens' correlation. However, standard self-attention suffers from quadratic memory and computation complexity, which arises from the $N \times N$ attention matrix, where $N$ is the sequence length. This problem is particularly significant during training, as it requires storing the attention matrix for gradient computation. Consequently, this significantly hinders the training of Transformer models with long sequences.

Recently, we have observed an increasing interest in training Transformer models with long sequences, especially when considering large language models \citep{scao2022bloom, zhang2022opt, chowdhery2022palm}. Various approaches address the quadratic memory issue in self-attention. One class of the methods is sparse-attention, which aims to perform only a subset of the attention computations while preserving the softmax function \citep{sparse_transformers_Child_2019, bigbird_Zaheer_2020}. Another approach is Linearized Attention (LA), which replaces the softmax with a product of two functions \citep{performer, linear_transformers2020Katharopoulos}. These methods reduce the computational and memory complexity of the attention mechanism while striving to maintain performance. One of LA's benefits is that it performs dense operations and does not require special HW or low-level implementation. However, despite their efficiency, LA methods often underperform compared to standard self-attention. Thus, understanding the reasons behind the superior performance of self-attention is crucial for designing an effective LA method.

In this paper, we propose a systematic way to develop an LA method with comparable performance to the Softmax Attention (SA). First, we define a holistic model of the SA and examine its characteristics. Then, we analyze the SA from three different perspectives, focusing on its statistical, informational, and algebraic properties. In particular, we characterize the probability distribution of the attention matrix and prove its log-normal nature. Moreover, we study the concentration behavior of the SA by analyzing its entropy and the spectral gap \citep{spectral_gap_Simon}. Based on the proposed model, we develop an LA method that emulates the distribution and concentration behavior of the SA, achieving comparable performance. Finally, we evaluate the effectiveness of our method on popular NLP benchmarks and compare it with other state-of-the-art methods. In summary, our contribution is as follows:
\begin{itemize}
    \item We conduct an in-depth analysis of self-attention, characterizing its statistical, informational, and algebraic properties.
    
    \item Develop tools suitable for studying the concentration ability of the attention based on the entropy and the spectral gap metrics. 
    
    \item Using the developed model and tools, we design Linear Log-Normal Attention (LLN Attention) with comparable performance to SA while requiring linear memory and computational complexity in the sequence length.
\end{itemize}

We have made the code of our method available for MindSpore\footnote{\href{https://gitee.com/ynahshan/linear-lognormal-attention-ms}{gitee.com/ynahshan/linear-lognormal-attention-ms}} and PyTorch\footnote{\href{https://github.com/ynahshan/linear-lognormal-attention}{github.com/ynahshan/linear-lognormal-attention}} frameworks.

\section{Background and related work}
In this section, we present a brief overview of the attention mechanism and various methods for efficient and linearized attention. We review the most relevant works in the field, classifying them into different types of attention methods such as sparse attention, low-rank projection, memory-based, kernel-based approximations, and clustering-based methods.

\subsection{Background on Self-Attention}
In the seminal study of \citep{attention-bahdanau}, the authors proposed the attention mechanism, which was subsequently incorporated into the Transformer model \citep{VaswaniSPUJGKP17}. Since then, attention has become a fundamental building block for many Transformer-based models.

Consider a sequence of $N$ tokens, where each token represented by $d$-dimensional query, key, and value vectors, denoted as $\{\mb{q}_i\}_{i=1}^N$, $\{\mb{k}_i\}_{i=1}^N$, and $\{\mb{v}_i\}_{i=1}^N$, respectively. The SA is defined as:

\begin{equation}
    \text{Attn}(\mb{q}_i, \{\mb{k}_1, \dots, \mb{k}_N\}, \{\mb{v}_1, \dots, \mb{v}_N\}) = \frac{\sum_{j=1}^N \kappa_{\rm exp}(\mb{q}_i, \mb{k}_j) \mb{v}^\top_j}{\sum_{l=1}^N \kappa_{\rm exp}(\mb{q}_i, \mb{k}_l)} 
    \label{eq:softmax_attention}
\end{equation}

where $\kappa_{\rm exp}$ is an exponential kernel used in the softmax function:
\begin{align}
    \kappa_{\rm exp}(\mb{q}_i,\mb{k}_j)={\rm e}^{\frac{\mb{q}_i^\top \mb{k}_j}{\sqrt{d}}}
\label{eq:softmax_attn_kernel}
\end{align}

The recent study by \citep{Transformers-RKBS} has examined SA from the perspective of the kernel method. Notably, the formulation of SA in \cref{eq:softmax_attention} can be seen as Nadaraya-Watson kernel regression \citep{nadaraya1964estimating}, where estimating some unknown function with joint distribution $p(\mb{k},\mb{v})$ and density $p(\mb{k})$ with a kernel \citep{Robustify-Han}. Moreover, as shown by \citep{Yao-Transformer-Dissection}, other popular kernels, such as polynomial or Radial Basis Function (RBF), can be used instead of the exponential kernel. However, the performance may vary depending on the type of the kernel. A kernel method perspective of the attention allows us to address the problem of attention linearization by using the connection between any kernel and its feature embedding function $\Phi$, described by Mercer's theorem \citep{mercer1909functions}:

\begin{align}
    \kappa(\mb{q}_i, \mb{k}_j) = \innerproduct{\Phi(\mb{q}_i)}{\Phi(\mb{k}_j)}
    \label{eq:mercer_kernel}
\end{align}

\subsection{Linearized Attention}
In recent years, several techniques have been proposed to address the quadratic cost associated with SA. Based on the taxonomy by \citep{long_short_zhu}, these techniques can be categorized into five types: i) sparse attention mechanisms with predefined patterns, including sliding window approaches such as Sparse Transformer \citep{sparse_transformers_Child_2019}, Axial Transformer \citep{axial_transformer_Ho_2019}, Blockwise Attention \citep{blockwise_Qiu_2019}, Longformer \citep{longformer_Beltagy_2020}, and BigBird \citep{bigbird_Zaheer_2020}, where some of these works \citep{kvt_eccv_2022} manage to improve model convergence due to noise reduction; ii) low-rank projection methods, including Linformer \citep{linformer_Wang_2020}, Synthesizer \citep{synthesizer_Tay_2020}, NystromFormer \citep{NystromFormer-2102-03902}, SkyFormer \citep{skyformer_Chen_2021}, and Cosformer \citep{cosformer-2202}; iii) memory-based methods, such as Set Transformer \citep{set_transformer_Lee_2018} and Compressive Transformers \citep{compressive_transformers_Rae_2019}; iv) kernel-based approximation of the attention matrix, including Performer \citep{performer}, Linear Transformers \citep{linear_transformers2020Katharopoulos}, and RFA \citep{rfa2021Peng}; and v) similarity and clustering methods, including Reformer \citep{reformer}, Routing Transformer \citep{routing_transformer_Roy_2020}, Sinkhorn Attention \citep{sinkhorn_attention_Tay_2020}, and Clustered Attention \citep{clustered_attention_Vyas_2020}.

Some of these methods combine multiple types of efficient attention mechanisms. For instance, \citep{long_short_zhu} suggested a combination of low-rank projection and local window attention. Similarly, \citep{linear_devil_Qin_2022} incorporate both kernel-based and block-wise attention in their approach. Our method also integrates kernel and block-wise techniques while suggesting a novel kernel approach that differs from that of \citep{linear_devil_Qin_2022}. By leveraging the benefits of multiple attention mechanisms, these techniques offer more efficient and accurate models for various NLP tasks.

Kernel-based attention requires selecting a feature embedding function $\Phi$ to compute the LA kernel \cref{eq:mercer_kernel}. Linearized attention can then be defined as:
\begin{equation}
\begin{split}
    \text{Attn}_{\text{lin}}(\mb{q}_i, \{\mb{k}_j\}_{j=1}^N, \{\mb{v}_j\}_{j=1}^N) = \frac{\sum_{j=1}^N \Phi_\mathcal{Q}(\mb{q}_i)^\top \Phi_\mathcal{K}(\mb{k}_j)}{\sum_{l=1}^N \Phi_\mathcal{Q}(\mb{q}_i)^\top\Phi_\mathcal{K}(\mb{k}_l)}\mb{v}^\top_j 
    = \frac{\Phi_\mathcal{Q}(\mb{q}_i)^\top\sum_{j=1}^N \Phi_\mathcal{K}(\mb{k}_j) \mb{v}^\top_j}{\Phi_\mathcal{Q}(\mb{q}_i)^\top\sum_{l=1}^N \Phi_\mathcal{K}(\mb{k}_l)} 
    \label{eq:linearized_attn}
\end{split}
\end{equation}
The choice of feature embedding function is crucial, as we demonstrate later in \cref{sec:design}. Different works suggest different types of embedding functions for this purpose. For example, Performer \citep{performer} uses an exponential function, Linear Transformers \citep{linear_transformers2020Katharopoulos} uses the ELU function, and RFA \citep{rfa2021Peng} uses trigonometric functions to approximate the Gaussian kernel with Fourier features. However, none of these works have analyzed the properties of the attention mechanism induced by these functions.
\section{Dissecting Softmax Attention}\label{sec:analysis}
In the previous section, we discussed the LA concept. Although this concept may seem straightforward, creating an LA mechanism that effectively handles complex tasks presents a challenging problem. Typically, the LA of the form in \cref{eq:linearized_attn} performs worse than the SA. To gain insight into the superiority of the SA, we conduct a thorough analysis of its properties. We start by characterizing the distribution of the attention matrix since it is a core element of the attention mechanism. Then, we study the connection between its entropy, spectral gap, and concentration ability of the self-attention.

We begin our analysis by formalizing a model based on the SA from \Cref{eq:softmax_attention}. Our model assumes that queries and keys approximately follow a Gaussian distribution. This assumption is reasonable due to the Central Limit Theorem (CLT) \citep{lee2018deep_gaussian} and accepted in literature for tractability purposes \citep{IoffeS15_BatchNorm, Banner_Quantization}. Moreover, let us assume the mean of queries and keys is approximately zero, which is a valid assumption due to the layer normalization presence in the Transformer models.

\begin{model_def}
Let $\mb{q}_i, \mb{k}_j \in \mathbb{R}^d$ be a Gaussian vectors, where elements $q_{i\ell} \sim \mathcal{N}(0,\sigma_q^2)$ and $k_{j \ell}\sim \mathcal{N}(0,\sigma_k^2)$, $\forall l$. Let \(a_{ij} = {\mb{q}_i^\top\mb{k}_j} \mathbin{/} \sqrt{d}\) the attention score of pair $i,j$, whose variance can be expressed as \(\sigma^2_{\rm sm} = \sigma^2_q\sigma^2_k + C_{\rm cross}\), where \(C_{\rm cross}\) is the cross-covariance of the squared queries and keys \citep{variance_of_products}. We define a temperature of the SA as:

\begin{equation}
    \tau_{\rm sm} = \frac{1}{\sigma_{\rm sm}} = \frac{1}{\sqrt{\sigma^2_q\sigma^2_k + C_{\rm cross}}}
    \label{eq:temperature_of_softmax_attention}
\end{equation}

Denote \(\Tilde{a}_{ij} = a_{ij} \mathbin{/} \sigma_{\rm sm}\) and let \(\mb{P}^{\text{(SM)}} \in \mathbb{R}^{N \times N}\) the SA matrix, where $N$ is sequence length such that:

\begin{equation}
    P_{ij}^{\text{(SM)}} = \frac{e^{\Tilde{a}_{ij} \mathbin{/} \tau_{\rm sm}}}{\sum_{l=1}^N e^{\Tilde{a}_{il} \mathbin{/} \tau_{\rm sm}}}
    \label{eq:softmax_attention_matrix}
\end{equation}
\end{model_def}
The form in \Cref{eq:softmax_attention_matrix} is significant as it demonstrates the connection between SA and implicit temperature parameter imposed by the variance of attention inputs. We can draw an analogy between the SA training and stochastic processes, where controlling the temperature allows balancing between exploration and exploitation. High temperature results in equal probabilities for all tokens (exploration), whereas low-temperature results in a high probability for one or few tokens, emphasizing it (exploitation of this particular token).

\subsection{Characterizing distribution of Softmax Attention}\label{sec:log-normal}
Let us now characterize the probability distribution of the SA. By analyzing its probability distribution, we can gain valuable insights into the behavior of the SA and reveal its statistical properties. In particular, the distribution of $\mb{P}^{\text{(SM)}}$ plays a crucial role in quantifying the variability of its entries. This variability is closely related to the concentration ability of the SA, a topic that we will explore in more detail in subsequent sections.

\begin{proposition}
\label{proposition:softmax_distribution}
Let $\mb{q}$ and $\mb{k}$ be Gaussian vectors, where $q_i\sim \mathcal{N}(0,\sigma_q^2)$ and $k_j\sim \mathcal{N}(0,\sigma_k^2)$, $\forall i,j$. Then, for moderate values of $\sigma_q^2, \sigma_k^2$ and large enough $N$ the distribution of $\mb{P}^{\text{(SM)}}$ can be approximated by a log-normal distribution with parameters \(\mu_\text{\rm sm} = -\ln N - \frac{1}{2}\sigma^2_{\rm sm}\) and \(\sigma^2_{\rm sm} = \sigma_q^2\sigma_k^2 + C_{\rm cross}\).
\end{proposition}

The key behind the proof is to approximate the denominator in \cref{eq:softmax_attention_matrix} with log-normal distribution by \citep{Fenton1960TheSO} theorem. Then, since the numerator is also log-normal by the CLT, the resulting ratio can be approximated by a log-normal distribution. It leads to the log-normal distribution of the $\mb{P}^{\text{(SM)}}$. The detailed proof is given in \Cref{section:softmax_distribution_appendix}.

The log-normal probability distribution of the SA matrix helps us understand the attention mechanism. The skewness of log-normal distribution emphasizes certain positions and enables concentration. The temperature parameter controls uncertainty, influencing the balance between exploration and exploitation during training.

\subsection{Analyzing Self-Attention through Markov Chain Perspective}
To delve deeper into the self-attention mechanism, we draw inspiration from the principles of Markov chains \citep{LevinPeresWilmer2006_MarkovChains}. Each row of the self-attention matrix represents the correlation between a specific token and all other tokens. These correlations closely resemble transition probabilities in classical Markov chains. The self-attention matrix continuously evolves during training, eventually converging to a final model.

\subsubsection{Entropy and Attention Concentration}

A crucial parameter in understanding such a stochastic system is its entropy, a metric commonly used to measure the uncertainty or randomness associated with the state transitions of a Markov chain. In the context of self-attention, entropy serves as a valuable tool to evaluate the concentration ability of the self-attention. We refer to this as Attention Concentration (AC), which essentially measures the model's ability to direct its focus toward specific tokens, thereby extracting relevant information from the input sequence. Previous studies \citep{Attention-Ghader, Attn_Structure_Belinkov} have proposed using entropy to measure the AC. Lower entropy indicates a greater focus on a few tokens, while higher entropy indicates more uniformly distributed attention. To formally define the entropy of the attention matrix $\mb{P}^{\text{(SM)}}$ we average the entropy of individual rows as following:
\begin{align}
    H(\mb{P}^{\text{(SM)}}) = - \frac{1}{N}\sum_{i=1}^N \sum_{j=1}^N P_{ij}^{\text{(SM)}} \log_2(P_{ij}^{\text{(SM)}})
    \label{eq:softmax_attention_entropy}
\end{align}

Note that by \cref{eq:softmax_attention_matrix}, the attention matrix can be represented in terms of its temperature. To further explore the connection between the AC and temperature, we present the following theorem, which characterizes the relationship between the entropy of the SA and its temperature:

\begin{theorem}
    The entropy in \cref{eq:softmax_attention_entropy} is monotonically increasing with temperature $\tau_{\rm sm}$.
\label{theorem:entropy_temperature_monotonicity}
\end{theorem}

To prove the theorem, we consider the derivative of the entropy with respect to the temperature and show it is always positive. For detailed proof, refer to \Cref{proof:entropy_temperature_monotonicity}. 

\begin{figure}
\centering
\begin{subfigure}[t]{0.325\textwidth}
 \centering
    \includegraphics[width=1.0\columnwidth]{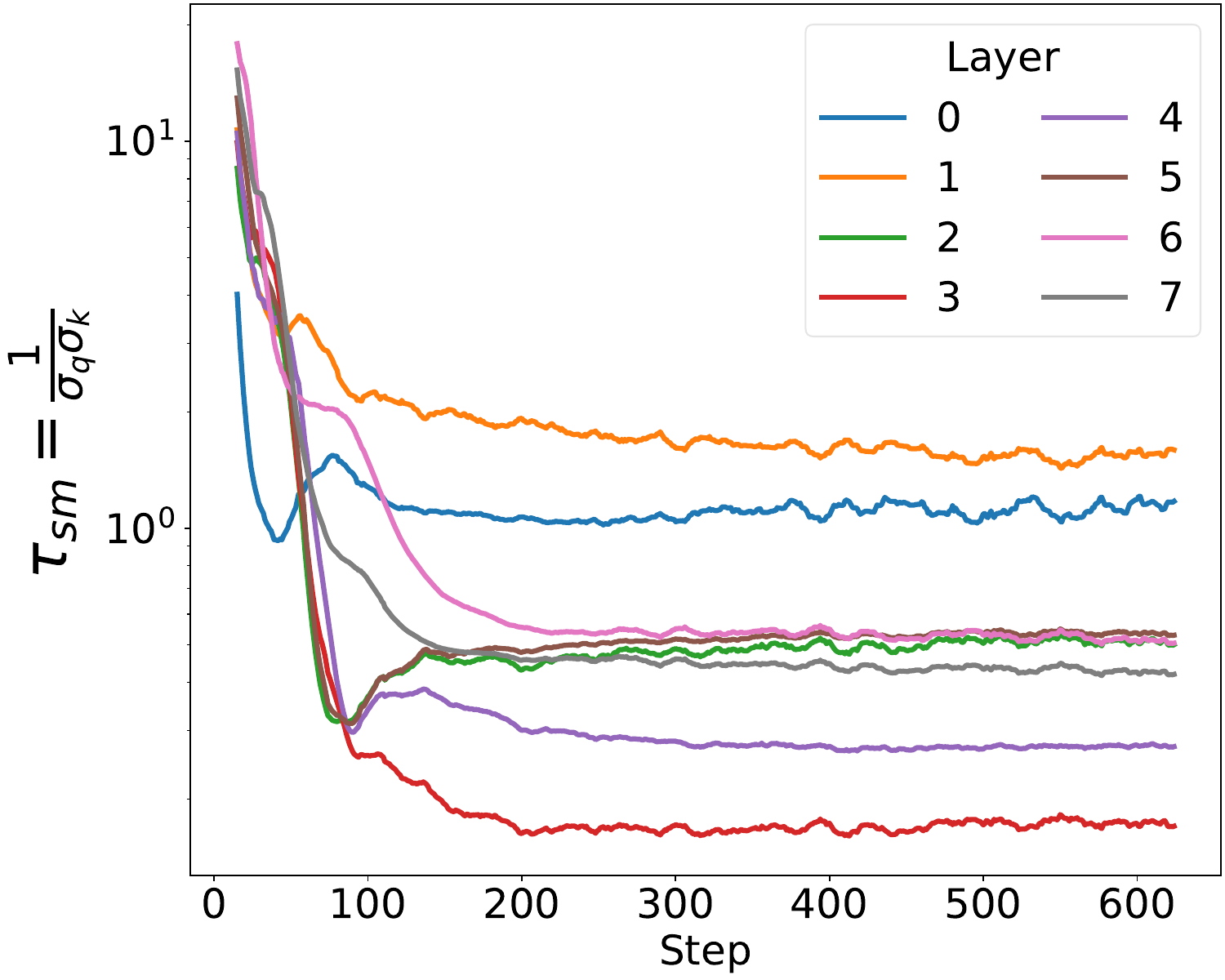}
    \caption{}
    \label{fig:roberta_temperature}
 \end{subfigure}
 \begin{subfigure}[t]{0.325\textwidth}
 \centering
    \includegraphics[width=1.00\columnwidth]
    {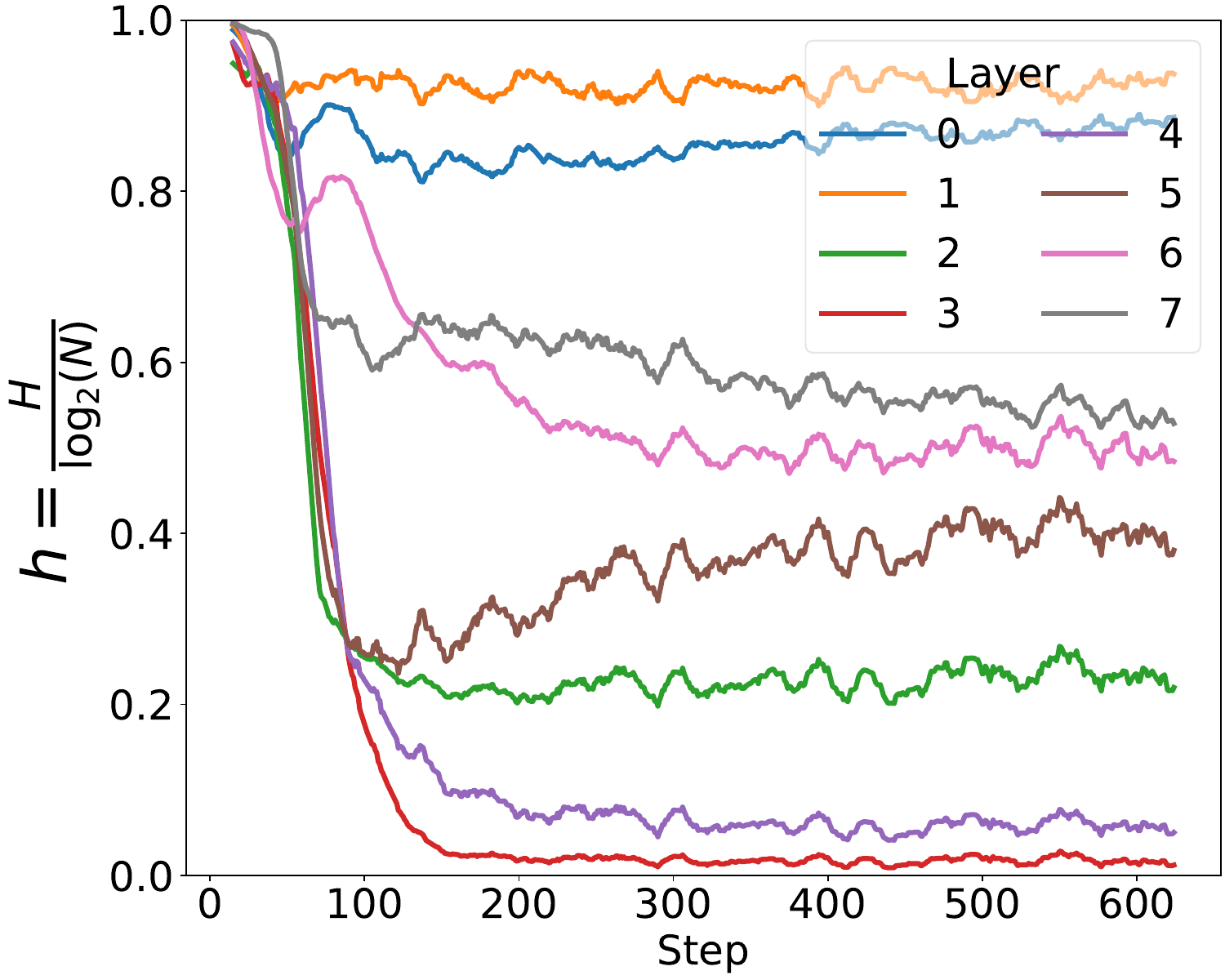}
    \caption{}
    \label{fig:roberta_entropy}
 \end{subfigure}
 \hfill
 \begin{subfigure}[t]{0.325\textwidth}
 \centering
    \includegraphics[width=1.0\columnwidth]{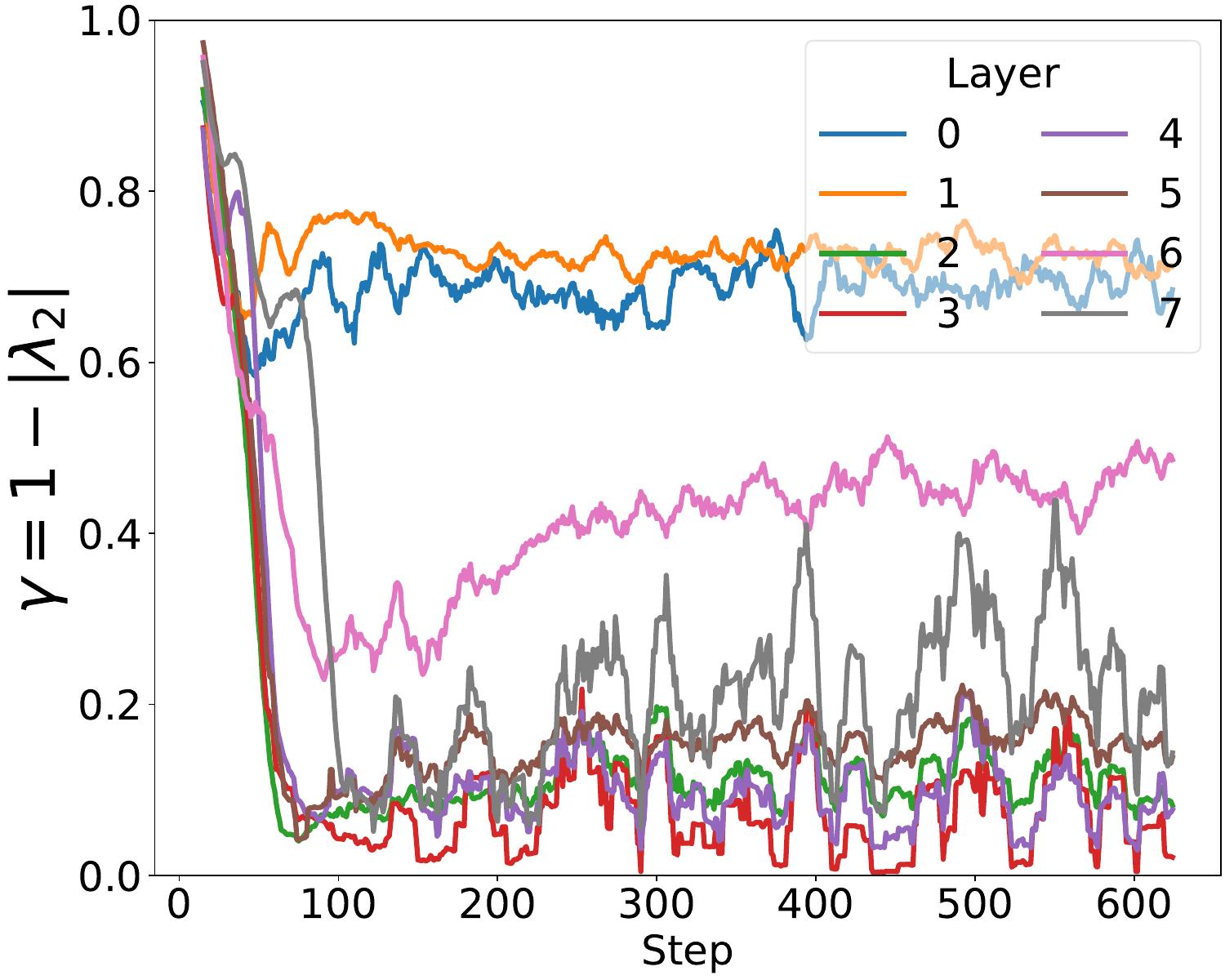}
    \caption{}
    \label{fig:roberta_spectral_gap}
 \end{subfigure}
 \caption{Temperature (left), entropy (center), and spectral gap (right) during training of the small RoBERTa model with a single head per layer in every training step (X-axis).}
 \label{fig:roberta_temp_entropy_spectral_gap}
\end{figure}

According to \Cref{theorem:entropy_temperature_monotonicity}, the entropy of the SA increases with the temperature, which controls the concentration of the SA. Essentially, a higher temperature results in a more dispersed distribution of attention. Conversely, a lower temperature makes it easier to focus on specific tokens. Moreover, the temperature controls the exploration (higher entropy) and exploitation (lower entropy) of the states within the chain.
\Cref{fig:roberta_temperature} shows how temperature decreases during training, resulting in a more confident state (lower entropy) \Cref{fig:roberta_entropy}. Notably, while the first layers of the model retain high entropy, allowing exploration, the entropy of the middle layers decreases and becomes approximately zero, leading to exploitation in those layers.

\subsubsection{Spectral Gap and Attention Concentration}
In the analysis of Markov chains, the spectral gap is a valuable metric to consider because it provides insights into the speed at which the chain reaches its stationary state \citep{spectral_gap_Simon}. In other words, it quantifies the rate of convergence, where the larger values of the spectral gap indicate a faster convergence process, while a smaller one suggests a slower one. When applied to attention mechanisms, the spectral gap can provide insights into the rate at which the attention mechanism focuses on specific elements within the input sequence. Including the spectral gap in our analysis allows us a more comprehensive understanding of the SA mechanism from an algebraic perspective.


The spectral gap measures the difference between the first and the second largest eigenvalues. Since the attention matrix is a \emph{stochastic matrix} \citep{meyer2000maa}, it follows from the Perron-Frobenius theorem \citep{Samelson1957_Perron-Frob} that its largest eigenvalue is $\lambda_1 = 1$. Therefore, the spectral gap is $\gamma = 1 - |\lambda_2|$, where $\lambda_2$ is the second largest eigenvalue. In \Cref{theorem:lambda2_var_max}, we establish a relationship between the variance of the attention matrix and the spectral gap.

\begin{theorem}
Let $\mb{P} \in \mathbb{R}^{N\times N}$ right stochastic matrix with eigenvalues $\lambda_1, \dots, \lambda_N$ ordered by their absolute values, where $\lambda_1=1 \ge |\lambda_2| \ge \dots \ge |\lambda_N|$. Let $\mb{\Bar{v}}_{\text{max}}$ be the major principal component of the centered version of $\mb{P}$. Then, \(\lambda_2^2 = \sigma^2_{\mb{\Bar{v}}_{\text{max}}}\), where $\sigma^2_{\mb{\Bar{v}}_{\text{max}}}$ represents the amount of variance in the direction specified by the major principal component $\mb{\Bar{v}}_{\text{max}}$.
\label{theorem:lambda2_var_max}
\end{theorem}

To prove this theorem, we deflate the $\lambda_1$ of the attention matrix and express the variance in the direction of the major principal component. The detailed proof is provided in \Cref{proof:lambda2_var_max}. 

\begin{theorem}
The variance of the attention matrix $\mb{P}^{\text{(SM)}}$ is monotonically decreasing with temperature $\tau_{\rm sm}$.
\label{theorem:var_vs_temperature}
\end{theorem}

To prove this theorem, we consider the derivative of the variance with respect to the temperature and show it is always negative. For detailed proof, refer to \Cref{theorem:var_temp_proof_appendix}. 

According to \Cref{theorem:lambda2_var_max}, the magnitude of variability in the direction of the major principal component \(\mb{\Bar{v}}_{\text{max}}\) is equal to $\lambda_2$, consequently, the spectral gap increases as the variability decreases. Together with the \Cref{theorem:var_vs_temperature}, we can conclude that the spectral gap increases with the temperature, similarly to the entropy. However, biasing the stochastic matrix towards a particular column also affects the variability. When $\mb{P}$ is biased toward a specific column, the variability within the columns decreases, resulting in a smaller value of $\lambda_2$ and a higher spectral gap, regardless of the temperature. Therefore, we can conclude that the spectral gap only increases with temperature when the attention matrix is unbiased. This phenomenon led us to refer to the spectral gap as a measure of \emph{Unbiased Attention Concentration}. \Cref{fig:roberta_spectral_gap} depicts the change in the spectral gap during training. In most layers, the spectral gap decreases during training, indicating improved AC. However, in some layers, the spectral gap increases while the temperature remains constant, suggesting that the attention matrix is biased. This observation justifies that the spectral gap carries additional information to entropy.



\section{Design of Linearized Attention}\label{sec:design}
In the previous section, we presented a holistic model of SA and conducted a thorough analysis of its properties. Specifically, we identified the log-normal distribution of the SA matrix. Additionally, we analyzed concentration behavior dictated by the temperature parameter. We can measure AC using the entropy (biased) and the spectral gap (unbiased) metrics. In this section, we design the LA method based on the defined model, which resembles similar characteristics and imitates SA behavior. In particular, our LA model should have log-normal distribution with similar moments. Moreover, it should emulate the concentration pattern of the SA by matching its entropy and spectral gap curves. As a result, we expect our LA method to achieve performance comparable to the SA.

\subsection{Linear Log-Normal Attention}
Designing LA according to \Cref{eq:linearized_attn} requires selecting a feature embedding function $\Phi$, a core element of this attention. The choice of this function has a crucial effect on the LA performance. According to our model, we start by satisfying the log-normality requirement, as most functions do not have this property. For example, the Rectified Linear Unit (ReLU) can not produce log-normal distribution as being almost linear. On the other hand, the exponential function induces log-normal distribution for Gaussian inputs, which justifies its selection as a feature embedding function $\Phi$. However, to match the concentration behavior of the SA, we must force the LA to produce similar entropy and spectral gap curves with respect to the temperature as in SA. To achieve this goal, we introduce additional parameters, which we tune to perform moment matching between the LA distribution and that of the SA.


Accordingly, let us denote by \(\Phi_\mathcal{Q}(\mb{q}) = e^{\alpha \mb{q}}\) and \(\Phi_\mathcal{K}(\mb{k}) = e^{\beta \mb{k}}\) the feature embedding functions, where $\alpha, \beta \in \mathbb{R}^+$ are hyper-parameters that must be carefully selected to ensure our LA closely approximates the SA. We define the Linear Log-Normal (LLN) Attention as:
\begin{equation}
    \text{Attn}_{\text{LLN}}(\mb{q}_i, \{\mb{k}_1, \dots, \mb{k}_N\}, \{\mb{v}_1, \dots, \mb{v}_N\}) = \frac{\sum_{j=1}^N e^{\alpha \mb{q}_i^\top} e^{\beta \mb{k}_j}}{\sum_{l=1}^N e^{\alpha \mb{q}_i^\top} e^{\beta \mb{k}_l}} \mb{v}^\top_j
    \label{eq:lln_attn}
\end{equation}
Where each entry of the LLN attention matrix can be expressed as:
\begin{equation}
    P_{ij}^\text{(LLN)} = \frac{e^{\alpha \mb{q}_i^\top} e^{\beta \mb{k}_j}}{\sum_{l=1}^N e^{\alpha \mb{q}_i^\top} e^{\beta \mb{k}_l}}
    \label{eq:lln_attention_matrix}
\end{equation}
Similarly to the analysis of the SA model, we assume zero mean of queries and keys. Then, to show that the LLN Attention matrix follows a log-normal distribution, we prove the following:
\begin{proposition}
\label{proposition:lln_distribution}
Let $\mb{q}$ and $\mb{k}$ be Gaussian vectors, where $q_i\sim \mathcal{N}(0,\sigma_q^2)$ and $k_j\sim \mathcal{N}(0,\sigma_k^2)$, $\forall i,j$. Then, for moderate values of $\sigma_q^2$ and $\sigma_k^2$, the distribution of $\mb{P}^{\text{(LLN)}}$ can be approximated by a log-normal distribution with variance \(\sigma_\text{lln}^2 = a\cdot(\alpha^2 \sigma^2_q + \beta^2 \sigma^2_k) + b\), where $a$ and $b$ are constants.
\end{proposition}

The main steps of the proof are approximating the numerator and denominator in \cref{eq:lln_attention_matrix} using the log-normal distribution, following the theorem by \citep{Fenton1960TheSO}. Then, split the analysis into three cases to express the variance, as suggested by \citep{Romeo_2003_lognormal_sum}. The detailed proof is given in \Cref{proposition:lln_distribution_proof}. 

Further, we have to ensure the concentration behavior of the LLN Attention is similar to that of the SA. To that end, it is necessary to determine appropriate values for the hyperparameters $\alpha$ and $\beta$. In the following, we estimate these parameters by performing moment matching to the distribution of the SA. Since the log-normal distribution is parameterized only by the first and second moments, we can align the LLN Attention distribution with the SA by ensuring equivalence of the first two moments.

\begin{figure}
\centering
 \begin{subfigure}[t]{0.4\textwidth}
 \centering
    \includegraphics[width=1.0\columnwidth]{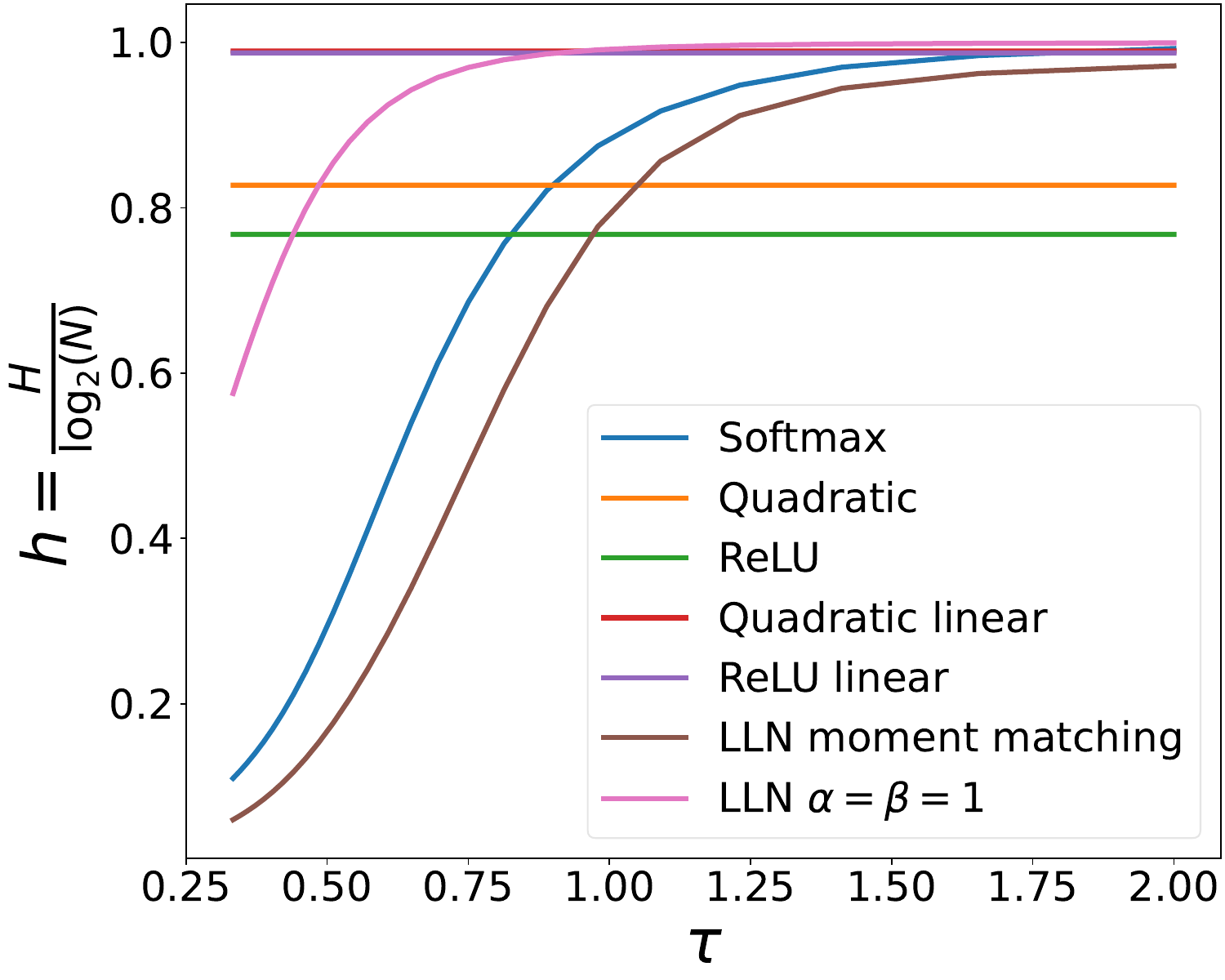}
 \end{subfigure}
 \hfill
 \begin{subfigure}[t]{0.4\textwidth}
 \centering
    \includegraphics[width=1.00\columnwidth]{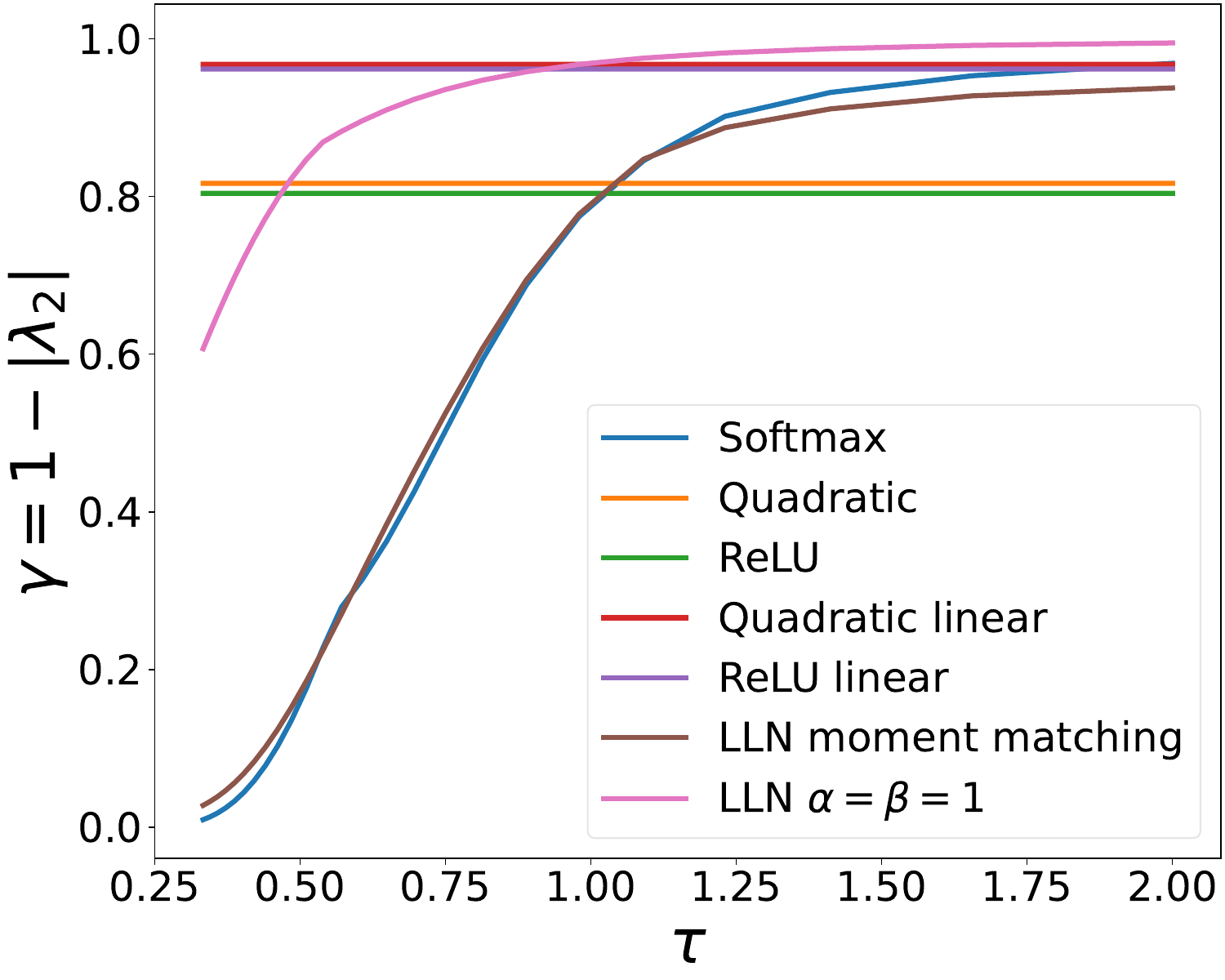}
 \end{subfigure}
\caption{Comparison of entropy (left) and spectral gap (right) for various types of attention kernels. The figure shows that the entropy and spectral gap of the LLN Attention with the moment matching is similar to those of the SA.}
\label{fig:various_kernels_spectral_gap_entropy}
\end{figure}

Interestingly, \Cref{proposition:lln_distribution} implies linear dependency between the variance of queries and keys and $\sigma_\text{lln}^2$. This linear connection facilitates the calculation of constants $a$ and $b$. It allows the application of linear interpolation on the randomly generated Gaussian samples $\mb{q}$ and $\mb{k}$ to perform the moment matching between LLN and SA. We provide a detailed description of the technical aspects of this moment-matching technique in \Cref{sec:moment_matching}.

Finally, by requiring $\sigma_\text{lln} = \sigma_\text{sm}$ and expressing it in terms of $\alpha$ and $\beta$, we can determine $a$ and $b$ parameters. We point out that there is no closed analytical solution for which both the mean and variance of LLN and SA align. Yet, since the concentration is mostly affected by the variance of the attention matrix, we only match the variances of the LLN and SA. Further,  to simplify the solution, we also let $\alpha^2 \sigma_q^2 = \beta^2 \sigma_k^2 = \frac{1}{2} \tilde{\sigma}^2$. Hence, we obtain the following:
\begin{equation}
    \alpha = \frac{\tilde{\sigma}}{\sqrt{2}\sigma_q}; \quad \beta = \frac{\tilde{\sigma}}{\sqrt{2}\sigma_k}; \quad \tilde{\sigma} = \sqrt{\frac{1}{a} (\sigma^2_q \sigma^2_k - b)}
    \label{eq:lln_alpha_beta}
\end{equation}
A detailed derivation of \Cref{eq:lln_alpha_beta} is given in \Cref{sec:moment_matching}.
Note that, like in the SA, we can introduce a temperature parameter of the LLN Attention that controls the concentration. Specifically, let us define the temperature of the LLN Attention to be:
\begin{align}
    \tau_{\rm lln} = \frac{1}{\sqrt{a\cdot(\alpha^2 \sigma^2_q + \beta^2 \sigma^2_k) + b}}
    \label{eq:lln_attn_temperature}
\end{align}
In  \Cref{fig:various_kernels_spectral_gap_entropy}, we demonstrate that the moment matching is essential to align the entropy and the spectral gap of the LLN Attention with those of the SA to achieve the required concentration. Moreover, other popular kernels, such as quadratic, ReLU, and their linear counterparts, are indifferent to the temperature, which may result in poor concentration and potentially degraded performance. In conclusion, LLN Attention satisfies the desired log-normal distribution property and concentration behavior required by the SA model. Therefore, it should achieve comparable results to the SA.

\subsection{The overall architecture}
\begin{wrapfigure}{r}{0.5\textwidth}
  \centering
    \includegraphics[width=0.4\textwidth]{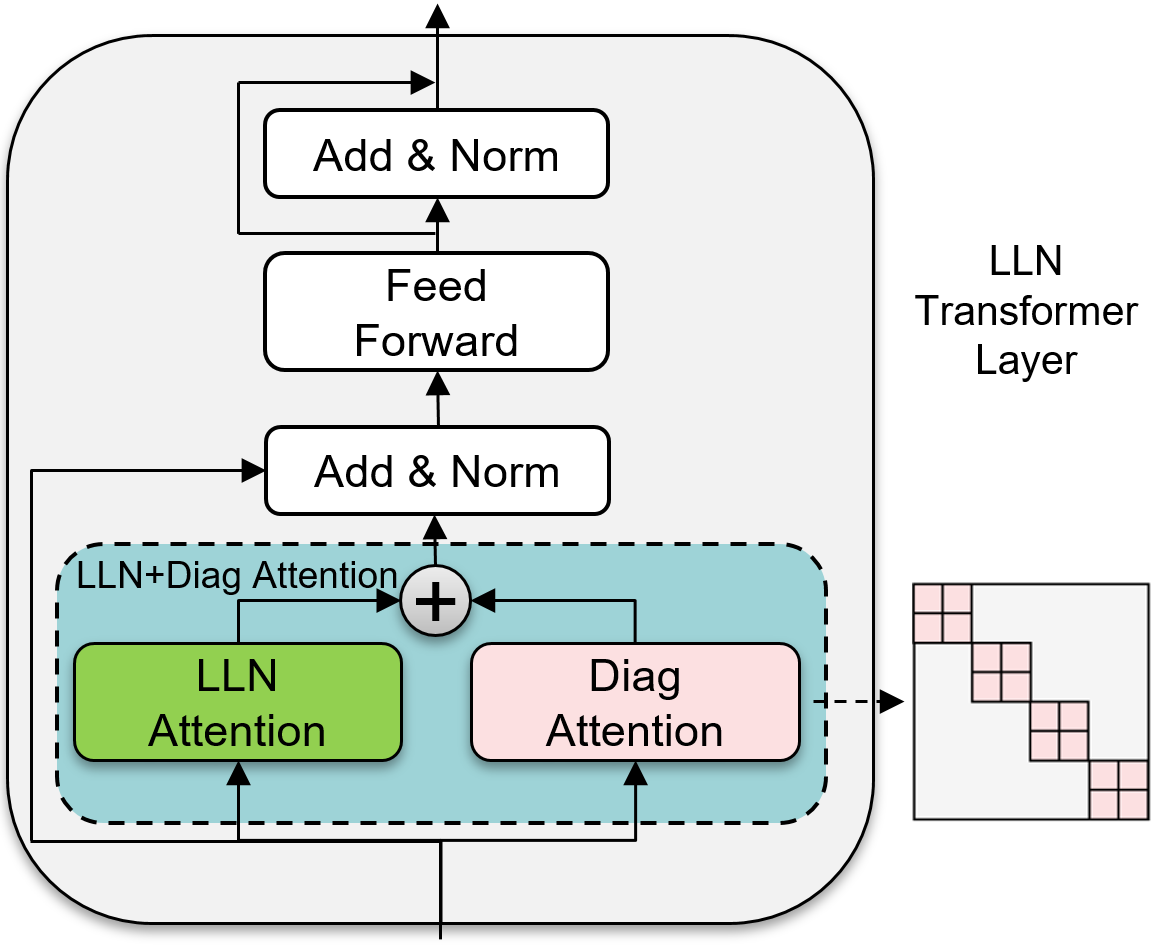}
  \caption{LLN Transformer layer architecture.}
  \label{fig:lln_transformer_layer}
\end{wrapfigure}
In this section, we present the experimental results of the LLN Attention method on NLP tasks, while more experiments on Image Classification and LRA\citep{lra-google} benchmark available in the \Cref{sec.appendinx_experiments}. 

The LLN Attention effectively scales to long sequences while maintaining high concentration, allowing capturing long-range interactions. However, for short-range connections, it may be less effective. Recently, a study by \citep{linear_devil_Qin_2022} emphasized the attention dilution issue of LA methods. Specifically, LA may overlook neighboring structures, leading to the "dilution" of short-range interactions. To address this issue, the authors proposed a hybrid approach that combines LA with block-diagonal attention, which retains the $O(N)$ memory and computational complexity of LA. This block-diagonal attention is a regular SA applied on smaller pieces of the input, computing only the diagonal of the original attention matrix. Such block-diagonal attention can not scale to longer sequences, but it is useful to improve the performance of the LA method.

We incorporate this technique into LLN Attention, combining the LLN and block-diagonal attention into a unified layer by averaging the outputs of both components \Cref{fig:lln_transformer_layer}. While the block-diagonal mechanism effectively captures short-range interactions within its confined block scope, LLN excels in catching broader, long-range connections. This combined approach enhances the performance of LLN Attention and stabilizes training by reducing the magnitude of the gradients \citep{linear_devil_Qin_2022}.

\section{Experiments}

We first pre-train the bidirectional RoBERTa encoder model \citep{Roberta_robustify} using LLN Attention on the WikiText-103 corpus \citep{Wikitext-103}. During pre-training, we monitor the convergence of the model and compare its performance to the SA model. In \Cref{sec.appendinx_experiments_roberta}, we show that the loss of the LLN Attention closely follows the loss of the SA, indicating similar convergence behavior.

Next, to evaluate the performance of LLN Attention on downstream tasks, we fine-tune our pre-trained model on several NLP tasks from the General Language Understanding Evaluation (GLUE) dataset \citep{GLUE_benchmark}. These tasks include Multi-Genre Natural Language Inference (MNLI), Question-answering Natural Language Inference (QNLI), Quora Question Pairs (QQP), and Stanford Sentiment Treebank (SST-2). For all our experiments, we use the Fairseq framework \citep{fairseq_facebook} with the default configuration and hyperparameters of the RoBERTa-base model.\footnote{https://github.com/facebookresearch/fairseq/blob/main/examples/roberta/README.md}

\Cref{tab:table_glue} provides a detailed comparison of the accuracy achieved by each method on each task. The LLN Attention method outperforms the other LA methods with an average accuracy of 86.9\%. These results confirm the superior capability of LLN Attention in achieving competitive performance with SA on a range of NLP tasks.
\begin{table}[]
\centering
\begin{tabular}{l|l|l|l|l|l|l}
\hline
\textbf{Method} & MNLI & QNLI & QQP & SST-2 & Avg $\uparrow$ \\
\hline
\hline
SA baseline \citep{attention-bahdanau} & 80.3 & 87.2 & 89.9 & 90.6 & 87.0 \\
\hline
Reformer \citep{reformer} & 35.4  & - & 63.2 & 50.9 & 49.8 \\
Performer \citep{performer} & 58.8 & 63.4 & 79.1 & 81.4 & 70.6 \\
ELU \citep{linear_transformers2020Katharopoulos} & 74.8 & 82.5 & 86.9 & 87.2 & 82.8 \\
Longformer \citep{longformer_Beltagy_2020} & 77.2 & - & 85.5 & 88.6 & 83.7 \\
Transformer LS \citep{long_short_zhu} & 77.0 & 84.8 & 86.8 & 90.2 & 84.7 \\
TNN \citep{qin2023toeplitz} & 76.72 & 85.06 & 88.3 & 90.6 & 85.17 \\
T2 \citep{linear_devil_Qin_2022} & 77.28 & 85.39 & 88.56 & 90.71 & 85.48 \\
CosFormer \citep{cosformer-2202} & 76.7 & - & 89.2 & 91 & 85.6 \\
T1 \citep{linear_devil_Qin_2022} & 79.06 & 87.0 & 88.61 & 91.17 & 86.46 \\
Flash \citep{hua2022transformer-flash} & 79.45 & \textbf{87.1} & 88.83 & 90.71 & 86.52 \\
Nystr\"{o}mformer$^\ast$ \citep{NystromFormer-2102-03902} & 80.9\small{(-1.5)} & 88.7\small{(-1.6)} & 86.3\small{(-1.)} & 91.4\small{(+1.4)} & 86.8\small{(-0.7)} \\
\hline
LLN Attention (Ours) & 77.0 & 85.1 & 88.9 & 90.6 & 85.4 \\
LLN+Diag Attention (Ours) & \textbf{80.0} & 86.5 & \textbf{89.7} & \textbf{91.6} & \textbf{86.9} \\
\hline
\end{tabular}
\caption{Accuracy achieved by various LA methods on multiple NLP tasks from the GLUE dataset, including MNLI, QNLI, QQP, and SST-2. The results of Nystr\"{o}mformer are given in \citep{NystromFormer-2102-03902}, while the results of the rest are given in \citep{linear_devil_Qin_2022}. Note that for methods marked with $^\ast$, which have a different baseline in the original paper, we also provide an accuracy drop in ().}
\label{tab:table_glue}
\vspace{-\baselineskip}
\end{table}

\subsection{Speed and memory consumption}
In this section, we evaluate the training time and memory usage of the LLN Attention, comparing it to the SA and Nystr\"{o}mformer, which outperform most of the LA methods available. In the comparison, we used the RoBERTa-base model with a batch size of one and performed all measurements on a commodity GPU.
The results in \Cref{tab:table_mem_perf} confirm that LLN Attention scales linearly with sequence length, as expected, and can handle at least four times longer sequences than SA. Moreover, the LLN Attention method requires nearly the same amount of memory as Nystr\"{o}mformer, with Diag Attention adding only a 10\% memory overhead to the LLN Attention. Notably, both LLN and LLN+Diag Attention demonstrate superior speed compared to Nystr\"{o}mformer.

\begin{table}
\centering
\begin{tabular}{clcccccc} \hline  &           & \multicolumn{6}{c}{sequence length}   \\ \hline
\multicolumn{1}{|l|}{}                        & \multicolumn{1}{l|}{Method}            & \multicolumn{1}{l|}{512}  & \multicolumn{1}{l|}{1024} & \multicolumn{1}{l|}{2048} & \multicolumn{1}{l|}{4096} & \multicolumn{1}{l|}{8192} & \multicolumn{1}{l|}{16384} \\ \hline\hline

\multicolumn{1}{|c|}{\multirow{4}{*}{Memory [GB]}} & \multicolumn{1}{l|}{Softmax Attention} & \multicolumn{1}{l|}{4.}   & \multicolumn{1}{l|}{5.5}  & \multicolumn{1}{l|}{12.6} & \multicolumn{1}{l|}{32.1} & \multicolumn{1}{l|}{OOM}  & \multicolumn{1}{l|}{OOM}   \\ \cline{2-8} 
\multicolumn{1}{|c|}{}                        & \multicolumn{1}{l|}{Nystr\"{o}mformer} & \multicolumn{1}{l|}{4.}   & \multicolumn{1}{l|}{4.5}  & \multicolumn{1}{l|}{5.5}  & \multicolumn{1}{l|}{7.3}  & \multicolumn{1}{l|}{11.6} & \multicolumn{1}{l|}{19.1}  \\ \cline{2-8} 
\multicolumn{1}{|c|}{}                        & \multicolumn{1}{l|}{LLN Attention}     & \multicolumn{1}{l|}{4.1}  & \multicolumn{1}{l|}{4.4}  & \multicolumn{1}{l|}{5.7}  & \multicolumn{1}{l|}{7.5}  & \multicolumn{1}{l|}{12.} & \multicolumn{1}{l|}{20.1}    \\  \cline{2-8} 
\multicolumn{1}{|c|}{}                        & \multicolumn{1}{l|}{LLN+Diag Attention}     & \multicolumn{1}{l|}{4.1}  & \multicolumn{1}{l|}{4.6}  & \multicolumn{1}{l|}{6.1}  & \multicolumn{1}{l|}{8.1}  & \multicolumn{1}{l|}{13.4} & \multicolumn{1}{l|}{23.}    \\ 
\hline\hline

\multicolumn{1}{|l|}{\multirow{4}{*}{Time [sec/it]}}   & \multicolumn{1}{l|}{Softmax Attention} & \multicolumn{1}{l|}{0.95} & \multicolumn{1}{l|}{1.05} & \multicolumn{1}{l|}{2.4}  & \multicolumn{1}{l|}{6.8}  & \multicolumn{1}{l|}{OOM}  & \multicolumn{1}{l|}{OOM}   \\ \cline{2-8} 
\multicolumn{1}{|l|}{}                        & \multicolumn{1}{l|}{Nystr\"{o}mformer} & \multicolumn{1}{l|}{1.8}  & \multicolumn{1}{l|}{1.9}  & \multicolumn{1}{l|}{2.6}  & \multicolumn{1}{l|}{4.7}  & \multicolumn{1}{l|}{8.8}  & \multicolumn{1}{l|}{16.7}  \\ \cline{2-8} 
\multicolumn{1}{|l|}{}                        & \multicolumn{1}{l|}{LLN Attention}     & \multicolumn{1}{l|}{1.}  & \multicolumn{1}{l|}{1.05}  & \multicolumn{1}{l|}{1.6}  & \multicolumn{1}{l|}{3.2}  & \multicolumn{1}{l|}{6.1}  & \multicolumn{1}{l|}{11.8}  \\ \cline{2-8} 
\multicolumn{1}{|l|}{}                        & \multicolumn{1}{l|}{LLN+Diag Attention}     & \multicolumn{1}{l|}{1.2}  & \multicolumn{1}{l|}{1.3}  & \multicolumn{1}{l|}{1.9}  & \multicolumn{1}{l|}{3.6}  & \multicolumn{1}{l|}{6.9}  & \multicolumn{1}{l|}{13.3}  \\ 
\hline
\end{tabular}
\caption{Memory usage and training time per iteration of SA, Nyströmformer,  LLN, and LLN+Diag on RoBERTa model, with varying sequence lengths. "OOM" indicates an "Out Of Memory" error.}
\label{tab:table_mem_perf}
\vspace{-\baselineskip}
\end{table}
\newpage
\section{Conclusion}
In this paper, we introduced a novel LLN Attention method that incorporates the essential properties of the SA, such as the log-normal distribution of the attention matrix and its concentration behavior, while offering linear time and memory complexity. Our approach includes a moment-matching technique to match the attention matrix's log-normal distribution with that of the SA, resulting in improved attention concentration and model performance. In addition, we conducted a comprehensive analysis of the SA, characterizing its distribution and suggesting entropy and the spectral gap metrics for attention concentration analysis. To the best of our knowledge, this is the first work to study self-attention from this perspective. Finally, our experimental results demonstrated that LLN Attention outperforms many existing LA methods on several NLP tasks, demonstrating its competitiveness and potential to enhance attention performance on long sequences. Overall, our contribution provides a foundation for future research and improvements in attention mechanisms.

\section*{Acknowledgements}
We extend our gratitude to Dr. Eliezer Levy, Dror Mizrachi, Dr. Su Teng, Wang ShengNan, and Dror Meirovich for their valuable support and fruitful discussions.

\newpage


\begin{thebibliography}{60}
\providecommand{\natexlab}[1]{#1}
\providecommand{\url}[1]{\texttt{#1}}
\expandafter\ifx\csname urlstyle\endcsname\relax
  \providecommand{\doi}[1]{doi: #1}\else
  \providecommand{\doi}{doi: \begingroup \urlstyle{rm}\Url}\fi

\bibitem[Vaswani et~al.(2017)Vaswani, Shazeer, Parmar, Uszkoreit, Jones, Gomez,
  Kaiser, and Polosukhin]{VaswaniSPUJGKP17}
Ashish Vaswani, Noam Shazeer, Niki Parmar, Jakob Uszkoreit, Llion Jones,
  Aidan~N. Gomez, Lukasz Kaiser, and Illia Polosukhin.
\newblock Attention is all you need, 2017.
\newblock URL \url{http://arxiv.org/abs/1706.03762}.

\bibitem[Brown et~al.(2020)Brown, Mann, Ryder, Subbiah, Kaplan, Dhariwal,
  Neelakantan, Shyam, Sastry, Askell, et~al.]{brown2020language}
Tom Brown, Benjamin Mann, Nick Ryder, Melanie Subbiah, Jared~D Kaplan, Prafulla
  Dhariwal, Arvind Neelakantan, Pranav Shyam, Girish Sastry, Amanda Askell,
  et~al.
\newblock Language models are few-shot learners.
\newblock \emph{Advances in neural information processing systems},
  33:\penalty0 1877--1901, 2020.

\bibitem[Devlin et~al.(2018)Devlin, Chang, Lee, and Toutanova]{Bert-2018}
Jacob Devlin, Ming{-}Wei Chang, Kenton Lee, and Kristina Toutanova.
\newblock {BERT:} pre-training of deep bidirectional transformers for language
  understanding, 2018.
\newblock URL \url{http://arxiv.org/abs/1810.04805}.

\bibitem[Dosovitskiy et~al.(2020)Dosovitskiy, Beyer, Kolesnikov, Weissenborn,
  Zhai, Unterthiner, Dehghani, Minderer, Heigold, Gelly, Uszkoreit, and
  Houlsby]{Vision-16-16}
Alexey Dosovitskiy, Lucas Beyer, Alexander Kolesnikov, Dirk Weissenborn,
  Xiaohua Zhai, Thomas Unterthiner, Mostafa Dehghani, Matthias Minderer, Georg
  Heigold, Sylvain Gelly, Jakob Uszkoreit, and Neil Houlsby.
\newblock An image is worth 16x16 words: Transformers for image recognition at
  scale, 2020.
\newblock URL \url{https://arxiv.org/abs/2010.11929}.

\bibitem[Chen et~al.(2018)Chen, Firat, Bapna, Johnson, Macherey, Foster, Jones,
  Parmar, Schuster, Chen, Wu, and Hughes]{NMT_best_of_words}
Mia~Xu Chen, Orhan Firat, Ankur Bapna, Melvin Johnson, Wolfgang Macherey,
  George~F. Foster, Llion Jones, Niki Parmar, Mike Schuster, Zhifeng Chen,
  Yonghui Wu, and Macduff Hughes.
\newblock The best of both worlds: Combining recent advances in neural machine
  translation, 2018.
\newblock URL \url{http://arxiv.org/abs/1804.09849}.

\bibitem[Zhang et~al.(2019)Zhang, Wei, and Zhou]{hilbert-doc-sum}
Xingxing Zhang, Furu Wei, and Ming Zhou.
\newblock {HIBERT:} document level pre-training of hierarchical bidirectional
  transformers for document summarization, 2019.
\newblock URL \url{http://arxiv.org/abs/1905.06566}.

\bibitem[Pilault et~al.(2020)Pilault, Li, Subramanian, and
  Pal]{pilault-etal-2020-extractive}
Jonathan Pilault, Raymond Li, Sandeep Subramanian, and Chris Pal.
\newblock On extractive and abstractive neural document summarization with
  transformer language models, November 2020.
\newblock URL \url{https://aclanthology.org/2020.emnlp-main.748}.

\bibitem[Bahdanau et~al.(2015)Bahdanau, Cho, and Bengio]{attention-bahdanau}
Dzmitry Bahdanau, {Kyung Hyun} Cho, and Yoshua Bengio.
\newblock Neural machine translation by jointly learning to align and
  translate, January 2015.
\newblock 3rd International Conference on Learning Representations, ICLR 2015 ;
  Conference date: 07-05-2015 Through 09-05-2015.

\bibitem[Scao et~al.(2022)Scao, Fan, Akiki, Pavlick, Ili{\'c}, Hesslow,
  Castagn{\'e}, Luccioni, Yvon, Gall{\'e}, et~al.]{scao2022bloom}
Teven~Le Scao, Angela Fan, Christopher Akiki, Ellie Pavlick, Suzana Ili{\'c},
  Daniel Hesslow, Roman Castagn{\'e}, Alexandra~Sasha Luccioni, Fran{\c{c}}ois
  Yvon, Matthias Gall{\'e}, et~al.
\newblock Bloom: A 176b-parameter open-access multilingual language model.
\newblock \emph{arXiv preprint arXiv:2211.05100}, 2022.

\bibitem[Zhang et~al.(2022)Zhang, Roller, Goyal, Artetxe, Chen, Chen, Dewan,
  Diab, Li, Lin, et~al.]{zhang2022opt}
Susan Zhang, Stephen Roller, Naman Goyal, Mikel Artetxe, Moya Chen, Shuohui
  Chen, Christopher Dewan, Mona Diab, Xian Li, Xi~Victoria Lin, et~al.
\newblock Opt: Open pre-trained transformer language models.
\newblock \emph{arXiv preprint arXiv:2205.01068}, 2022.

\bibitem[Chowdhery et~al.(2022)Chowdhery, Narang, Devlin, Bosma, Mishra,
  Roberts, Barham, Chung, Sutton, Gehrmann, et~al.]{chowdhery2022palm}
Aakanksha Chowdhery, Sharan Narang, Jacob Devlin, Maarten Bosma, Gaurav Mishra,
  Adam Roberts, Paul Barham, Hyung~Won Chung, Charles Sutton, Sebastian
  Gehrmann, et~al.
\newblock Palm: Scaling language modeling with pathways.
\newblock \emph{arXiv preprint arXiv:2204.02311}, 2022.

\bibitem[Child et~al.(2019)Child, Gray, Radford, and
  Sutskever]{sparse_transformers_Child_2019}
Rewon Child, Scott Gray, Alec Radford, and Ilya Sutskever.
\newblock Generating long sequences with sparse transformers, 2019.
\newblock URL \url{http://arxiv.org/abs/1904.10509}.

\bibitem[Zaheer et~al.(2020)Zaheer, Guruganesh, Dubey, Ainslie, Alberti,
  Onta{\~{n}}{\'{o}}n, Pham, Ravula, Wang, Yang, and
  Ahmed]{bigbird_Zaheer_2020}
Manzil Zaheer, Guru Guruganesh, Avinava Dubey, Joshua Ainslie, Chris Alberti,
  Santiago Onta{\~{n}}{\'{o}}n, Philip Pham, Anirudh Ravula, Qifan Wang,
  Li~Yang, and Amr Ahmed.
\newblock Big bird: Transformers for longer sequences, 2020.
\newblock URL \url{https://arxiv.org/abs/2007.14062}.

\bibitem[Choromanski et~al.(2020)Choromanski, Likhosherstov, Dohan, Song, Gane,
  Sarl{\'{o}}s, Hawkins, Davis, Mohiuddin, Kaiser, Belanger, Colwell, and
  Weller]{performer}
Krzysztof Choromanski, Valerii Likhosherstov, David Dohan, Xingyou Song,
  Andreea Gane, Tam{\'{a}}s Sarl{\'{o}}s, Peter Hawkins, Jared Davis, Afroz
  Mohiuddin, Lukasz Kaiser, David Belanger, Lucy~J. Colwell, and Adrian Weller.
\newblock Rethinking attention with performers, 2020.
\newblock URL \url{https://arxiv.org/abs/2009.14794}.

\bibitem[Katharopoulos et~al.(2020)Katharopoulos, Vyas, Pappas, and
  Fleuret]{linear_transformers2020Katharopoulos}
Angelos Katharopoulos, Apoorv Vyas, Nikolaos Pappas, and Fran{\c{c}}ois
  Fleuret.
\newblock Transformers are rnns: Fast autoregressive transformers with linear
  attention, 2020.
\newblock URL \url{https://arxiv.org/abs/2006.16236}.

\bibitem[Coste(2017)]{spectral_gap_Simon}
Simon Coste.
\newblock The spectral gap of sparse random digraphs, 2017.
\newblock URL \url{https://arxiv.org/abs/1708.00530}.

\bibitem[Wright and Gonzalez(2021)]{Transformers-RKBS}
Matthew~A. Wright and Joseph~E. Gonzalez.
\newblock Transformers are deep infinite-dimensional non-mercer binary kernel
  machines, 2021.
\newblock URL \url{https://arxiv.org/abs/2106.01506}.

\bibitem[Nadaraya(1964)]{nadaraya1964estimating}
Elizbar~A Nadaraya.
\newblock On estimating regression, 1964.

\bibitem[Han et~al.(2022)Han, Ren, Nguyen, Nguyen, Ghosh, and
  Ho]{Robustify-Han}
Xing Han, Tongzheng Ren, Tan~Minh Nguyen, Khai Nguyen, Joydeep Ghosh, and Nhat
  Ho.
\newblock Robustify transformers with robust kernel density estimation, 2022.
\newblock URL \url{https://arxiv.org/abs/2210.05794}.

\bibitem[Tsai et~al.(2019)Tsai, Bai, Yamada, Morency, and
  Salakhutdinov]{Yao-Transformer-Dissection}
Yao{-}Hung~Hubert Tsai, Shaojie Bai, Makoto Yamada, Louis{-}Philippe Morency,
  and Ruslan Salakhutdinov.
\newblock Transformer dissection: An unified understanding for transformer's
  attention via the lens of kernel, 2019.
\newblock URL \url{http://arxiv.org/abs/1908.11775}.

\bibitem[Mercer(1909)]{mercer1909functions}
J.~Mercer.
\newblock Functions of positive and negative type, and their connection with
  the theory of integral equations, 1909.

\bibitem[Zhu et~al.(2021)Zhu, Ping, Xiao, Shoeybi, Goldstein, Anandkumar, and
  Catanzaro]{long_short_zhu}
Chen Zhu, Wei Ping, Chaowei Xiao, Mohammad Shoeybi, Tom Goldstein, Anima
  Anandkumar, and Bryan Catanzaro.
\newblock Long-short transformer: Efficient transformers for language and
  vision, 2021.
\newblock URL \url{https://arxiv.org/abs/2107.02192}.

\bibitem[Ho et~al.(2019)Ho, Kalchbrenner, Weissenborn, and
  Salimans]{axial_transformer_Ho_2019}
Jonathan Ho, Nal Kalchbrenner, Dirk Weissenborn, and Tim Salimans.
\newblock Axial attention in multidimensional transformers, 2019.
\newblock URL \url{http://arxiv.org/abs/1912.12180}.

\bibitem[Qiu et~al.(2019)Qiu, Ma, Levy, Yih, Wang, and
  Tang]{blockwise_Qiu_2019}
Jiezhong Qiu, Hao Ma, Omer Levy, Scott~Wen{-}tau Yih, Sinong Wang, and Jie
  Tang.
\newblock Blockwise self-attention for long document understanding, 2019.
\newblock URL \url{http://arxiv.org/abs/1911.02972}.

\bibitem[Beltagy et~al.(2020)Beltagy, Peters, and
  Cohan]{longformer_Beltagy_2020}
Iz~Beltagy, Matthew~E. Peters, and Arman Cohan.
\newblock Longformer: The long-document transformer, 2020.
\newblock URL \url{https://arxiv.org/abs/2004.05150}.

\bibitem[Wang et~al.(2021)Wang, Wang, Wang, Lin, Chang, Xie, Li, and
  Jin]{kvt_eccv_2022}
Pichao Wang, Xue Wang, Fan Wang, Ming Lin, Shuning Chang, Wen Xie, Hao Li, and
  Rong Jin.
\newblock {KVT:} k-nn attention for boosting vision transformers.
\newblock \emph{CoRR}, abs/2106.00515, 2021.
\newblock URL \url{https://arxiv.org/abs/2106.00515}.

\bibitem[Wang et~al.(2020)Wang, Li, Khabsa, Fang, and Ma]{linformer_Wang_2020}
Sinong Wang, Belinda~Z. Li, Madian Khabsa, Han Fang, and Hao Ma.
\newblock Linformer: Self-attention with linear complexity, 2020.
\newblock URL \url{https://arxiv.org/abs/2006.04768}.

\bibitem[Tay et~al.(2020{\natexlab{a}})Tay, Bahri, Metzler, Juan, Zhao, and
  Zheng]{synthesizer_Tay_2020}
Yi~Tay, Dara Bahri, Donald Metzler, Da{-}Cheng Juan, Zhe Zhao, and Che Zheng.
\newblock Synthesizer: Rethinking self-attention in transformer models,
  2020{\natexlab{a}}.
\newblock URL \url{https://arxiv.org/abs/2005.00743}.

\bibitem[Xiong et~al.(2021)Xiong, Zeng, Chakraborty, Tan, Fung, Li, and
  Singh]{NystromFormer-2102-03902}
Yunyang Xiong, Zhanpeng Zeng, Rudrasis Chakraborty, Mingxing Tan, Glenn Fung,
  Yin Li, and Vikas Singh.
\newblock Nystr{\"{o}}mformer: {A} nystr{\"{o}}m-based algorithm for
  approximating self-attention, 2021.
\newblock URL \url{https://arxiv.org/abs/2102.03902}.

\bibitem[Chen et~al.(2021)Chen, Zeng, Ji, and Yang]{skyformer_Chen_2021}
Yifan Chen, Qi~Zeng, Heng Ji, and Yun Yang.
\newblock Skyformer: Remodel self-attention with gaussian kernel and
  nystr{\"{o}}m method, 2021.
\newblock URL \url{https://arxiv.org/abs/2111.00035}.

\bibitem[Qin et~al.(2022{\natexlab{a}})Qin, Sun, Deng, Li, Wei, Lv, Yan, Kong,
  and Zhong]{cosformer-2202}
Zhen Qin, Weixuan Sun, Hui Deng, Dongxu Li, Yunshen Wei, Baohong Lv, Junjie
  Yan, Lingpeng Kong, and Yiran Zhong.
\newblock cosformer: Rethinking softmax in attention, 2022{\natexlab{a}}.
\newblock URL \url{https://arxiv.org/abs/2202.08791}.

\bibitem[Lee et~al.(2018{\natexlab{a}})Lee, Lee, Kim, Kosiorek, Choi, and
  Teh]{set_transformer_Lee_2018}
Juho Lee, Yoonho Lee, Jungtaek Kim, Adam~R. Kosiorek, Seungjin Choi, and
  Yee~Whye Teh.
\newblock Set transformer, 2018{\natexlab{a}}.
\newblock URL \url{http://arxiv.org/abs/1810.00825}.

\bibitem[Rae et~al.(2019)Rae, Potapenko, Jayakumar, and
  Lillicrap]{compressive_transformers_Rae_2019}
Jack~W. Rae, Anna Potapenko, Siddhant~M. Jayakumar, and Timothy~P. Lillicrap.
\newblock Compressive transformers for long-range sequence modelling, 2019.
\newblock URL \url{http://arxiv.org/abs/1911.05507}.

\bibitem[Peng et~al.(2021)Peng, Pappas, Yogatama, Schwartz, Smith, and
  Kong]{rfa2021Peng}
Hao Peng, Nikolaos Pappas, Dani Yogatama, Roy Schwartz, Noah~A. Smith, and
  Lingpeng Kong.
\newblock Random feature attention, 2021.
\newblock URL \url{https://arxiv.org/abs/2103.02143}.

\bibitem[Kitaev et~al.(2020)Kitaev, Kaiser, and Levskaya]{reformer}
Nikita Kitaev, Lukasz Kaiser, and Anselm Levskaya.
\newblock Reformer: The efficient transformer, 2020.
\newblock URL \url{https://arxiv.org/abs/2001.04451}.

\bibitem[Roy et~al.(2020)Roy, Saffar, Vaswani, and
  Grangier]{routing_transformer_Roy_2020}
Aurko Roy, Mohammad Saffar, Ashish Vaswani, and David Grangier.
\newblock Efficient content-based sparse attention with routing transformers,
  2020.
\newblock URL \url{https://arxiv.org/abs/2003.05997}.

\bibitem[Tay et~al.(2020{\natexlab{b}})Tay, Bahri, Yang, Metzler, and
  Juan]{sinkhorn_attention_Tay_2020}
Yi~Tay, Dara Bahri, Liu Yang, Donald Metzler, and Da{-}Cheng Juan.
\newblock Sparse sinkhorn attention, 2020{\natexlab{b}}.
\newblock URL \url{https://arxiv.org/abs/2002.11296}.

\bibitem[Vyas et~al.(2020)Vyas, Katharopoulos, and
  Fleuret]{clustered_attention_Vyas_2020}
Apoorv Vyas, Angelos Katharopoulos, and Fran{\c{c}}ois Fleuret.
\newblock Fast transformers with clustered attention, 2020.
\newblock URL \url{https://arxiv.org/abs/2007.04825}.

\bibitem[Qin et~al.(2022{\natexlab{b}})Qin, Han, Sun, Li, Kong, Barnes, and
  Zhong]{linear_devil_Qin_2022}
Zhen Qin, XiaoDong Han, Weixuan Sun, Dongxu Li, Lingpeng Kong, Nick Barnes, and
  Yiran Zhong.
\newblock The devil in linear transformer, 2022{\natexlab{b}}.
\newblock URL \url{https://arxiv.org/abs/2210.10340}.

\bibitem[Lee et~al.(2018{\natexlab{b}})Lee, Bahri, Novak, Schoenholz,
  Pennington, and Sohl-Dickstein]{lee2018deep_gaussian}
Jaehoon Lee, Yasaman Bahri, Roman Novak, Samuel~S. Schoenholz, Jeffrey
  Pennington, and Jascha Sohl-Dickstein.
\newblock Deep neural networks as gaussian processes, 2018{\natexlab{b}}.

\bibitem[Ioffe and Szegedy(2015)]{IoffeS15_BatchNorm}
Sergey Ioffe and Christian Szegedy.
\newblock Batch normalization: Accelerating deep network training by reducing
  internal covariate shift.
\newblock \emph{CoRR}, abs/1502.03167, 2015.
\newblock URL \url{http://arxiv.org/abs/1502.03167}.

\bibitem[Banner et~al.(2018)Banner, Nahshan, Hoffer, and
  Soudry]{Banner_Quantization}
Ron Banner, Yury Nahshan, Elad Hoffer, and Daniel Soudry.
\newblock {ACIQ:} analytical clipping for integer quantization of neural
  networks.
\newblock \emph{CoRR}, abs/1810.05723, 2018.
\newblock URL \url{http://arxiv.org/abs/1810.05723}.

\bibitem[Goodman(1960)]{variance_of_products}
Leo~A. Goodman.
\newblock On the exact variance of products, 1960.
\newblock URL
  \url{https://www.tandfonline.com/doi/abs/10.1080/01621459.1960.10483369}.

\bibitem[Fenton(1960)]{Fenton1960TheSO}
Leslie~H. Fenton.
\newblock The sum of log-normal probability distributions in scatter
  transmission systems, 1960.

\bibitem[Levin et~al.(2006)Levin, Peres, and
  Wilmer]{LevinPeresWilmer2006_MarkovChains}
David~A. Levin, Yuval Peres, and Elizabeth~L. Wilmer.
\newblock \emph{{Markov chains and mixing times}}.
\newblock American Mathematical Society, 2006.
\newblock URL
  \url{http://scholar.google.com/scholar.bib?q=info:3wf9IU94tyMJ:scholar.google.com/&output=citation&hl=en&as_sdt=2000&ct=citation&cd=0}.

\bibitem[Ghader and Monz(2017)]{Attention-Ghader}
Hamidreza Ghader and Christof Monz.
\newblock What does attention in neural machine translation pay attention to?,
  2017.
\newblock URL \url{http://arxiv.org/abs/1710.03348}.

\bibitem[Vig and Belinkov(2019)]{Attn_Structure_Belinkov}
Jesse Vig and Yonatan Belinkov.
\newblock Analyzing the structure of attention in a transformer language model,
  2019.
\newblock URL \url{http://arxiv.org/abs/1906.04284}.

\bibitem[Meyer(2000)]{meyer2000maa}
C.D. Meyer.
\newblock \emph{{Matrix Analysis and Applied Linear Algebra}}.
\newblock Society for Industrial Mathematics, 2000.
\newblock URL \url{http://www.matrixanalysis.com/DownloadChapters.html}.

\bibitem[Samelson(1957)]{Samelson1957_Perron-Frob}
Hans Samelson.
\newblock On the perron-frobenius theorem., 1957.

\bibitem[Romeo et~al.(2003)Romeo, Costa, and Bardou]{Romeo_2003_lognormal_sum}
M.~Romeo, V.~Da Costa, and F.~Bardou.
\newblock Broad distribution effects in sums of lognormal random variables, apr
  2003.
\newblock URL \url{https://doi.org/10.1140%2Fepjb%2Fe2003-00131-6}.

\bibitem[Tay et~al.(2020{\natexlab{c}})Tay, Dehghani, Abnar, Shen, Bahri, Pham,
  Rao, Yang, Ruder, and Metzler]{lra-google}
Yi~Tay, Mostafa Dehghani, Samira Abnar, Yikang Shen, Dara Bahri, Philip Pham,
  Jinfeng Rao, Liu Yang, Sebastian Ruder, and Donald Metzler.
\newblock Long range arena: {A} benchmark for efficient transformers,
  2020{\natexlab{c}}.
\newblock URL \url{https://arxiv.org/abs/2011.04006}.

\bibitem[Liu et~al.(2019)Liu, Ott, Goyal, Du, Joshi, Chen, Levy, Lewis,
  Zettlemoyer, and Stoyanov]{Roberta_robustify}
Yinhan Liu, Myle Ott, Naman Goyal, Jingfei Du, Mandar Joshi, Danqi Chen, Omer
  Levy, Mike Lewis, Luke Zettlemoyer, and Veselin Stoyanov.
\newblock Roberta: {A} robustly optimized {BERT} pretraining approach, 2019.
\newblock URL \url{http://arxiv.org/abs/1907.11692}.

\bibitem[Merity et~al.(2018)Merity, Keskar, and Socher]{Wikitext-103}
Stephen Merity, Nitish~Shirish Keskar, and Richard Socher.
\newblock An analysis of neural language modeling at multiple scales, 2018.
\newblock URL \url{https://arxiv.org/abs/1803.08240}.

\bibitem[Wang et~al.(2018)Wang, Singh, Michael, Hill, Levy, and
  Bowman]{GLUE_benchmark}
Alex Wang, Amanpreet Singh, Julian Michael, Felix Hill, Omer Levy, and
  Samuel~R. Bowman.
\newblock {GLUE:} {A} multi-task benchmark and analysis platform for natural
  language understanding, 2018.
\newblock URL \url{http://arxiv.org/abs/1804.07461}.

\bibitem[Ott et~al.(2019)Ott, Edunov, Baevski, Fan, Gross, Ng, Grangier, and
  Auli]{fairseq_facebook}
Myle Ott, Sergey Edunov, Alexei Baevski, Angela Fan, Sam Gross, Nathan Ng,
  David Grangier, and Michael Auli.
\newblock fairseq: {A} fast, extensible toolkit for sequence modeling.
\newblock \emph{CoRR}, abs/1904.01038, 2019.
\newblock URL \url{http://arxiv.org/abs/1904.01038}.

\bibitem[Qin et~al.(2023)Qin, Han, Sun, He, Li, Li, Dai, Kong, and
  Zhong]{qin2023toeplitz}
Zhen Qin, Xiaodong Han, Weixuan Sun, Bowen He, Dong Li, Dongxu Li, Yuchao Dai,
  Lingpeng Kong, and Yiran Zhong.
\newblock Toeplitz neural network for sequence modeling.
\newblock In \emph{The Eleventh International Conference on Learning
  Representations}, 2023.
\newblock URL \url{https://openreview.net/forum?id=IxmWsm4xrua}.

\bibitem[Hua et~al.(2022)Hua, Dai, Liu, and Le]{hua2022transformer-flash}
Weizhe Hua, Zihang Dai, Hanxiao Liu, and Quoc~V. Le.
\newblock Transformer quality in linear time, 2022.

\bibitem[Jolliffe(2011)]{Jolliffe2011}
Ian Jolliffe.
\newblock Principal component analysis, 2011.
\newblock URL \url{https://doi.org/10.1007/978-3-642-04898-2_455}.

\bibitem[Weisstein(2003)]{weisstein2003normal}
Eric~W Weisstein.
\newblock Normal product distribution, 2003.

\bibitem[Touvron et~al.(2020)Touvron, Cord, Douze, Massa, Sablayrolles, and
  J{\'{e}}gou]{deit_touvron}
Hugo Touvron, Matthieu Cord, Matthijs Douze, Francisco Massa, Alexandre
  Sablayrolles, and Herv{\'{e}} J{\'{e}}gou.
\newblock Training data-efficient image transformers {\&} distillation through
  attention.
\newblock \emph{CoRR}, abs/2012.12877, 2020.
\newblock URL \url{https://arxiv.org/abs/2012.12877}.

\end{thebibliography}

\newpage
\appendix{}
\section{Appendix}
\subsection{Formulation of the self-attention (SA)}
One of the most widely used variants of the self-attention mechanism is the scaled dot-product attention \citep{VaswaniSPUJGKP17}. In this formulation, the input vectors $[\mb{x}_1, \dots, \mb{x}_N]^\top := \mb{X} \in \mathbb{R}^{N \times d}$ are projected into query, key, and value vectors as follows:
\begin{equation}
\begin{split}
 [\mb{q}_1, \dots, \mb{q}_N]^\top := \mb{Q} = \mb{X}\mb{W}_q^\top  \in \mathbb{R}^{N \times d} \\
 [\mb{k}_1, \dots, \mb{k}_N]^\top := \mb{K} = \mb{X}\mb{W}_k^\top  \in \mathbb{R}^{N \times d} \\
 [\mb{v}_1, \dots, \mb{v}_N]^\top := \mb{V} = \mb{X}\mb{W}_v^\top  \in \mathbb{R}^{N \times d}
\end{split}
\end{equation}
Here, $\mb{W}_q, \mb{W}_k, \mb{W}_v \in \mathbb{R}^{d \times d}$ are learnable parameter matrices. The attention function is then computed on the query, key, and value vectors using the following equation:

\begin{equation}
    \text{Attn}(\mb{q}_i, \{\mb{k}_1, \dots, \mb{k}_N\}, \{\mb{v}_1, \dots, \mb{v}_N\}) = \sum_{j=1}^N \text{softmax}\left(\frac{\mb{q}_i^\top \mb{k}_j}{\sqrt{d}}\right) \mb{v}^\top_j 
    \label{eq:softmax_attention_appendix}
\end{equation}

The dot-product term is scaled by the factor $\frac{1}{\sqrt{d}}$ to ensure the stability of computations. This scaling factor accounts for the variance of the dot-product $\mb{q}_i^\top \mb{k}_j$, which grows with the dimensionality $d$. In this paper, we refer to this scaled dot-product attention as "softmax attention" or "standard attention," which is defined in \cref{eq:softmax_attention}.

\subsection{Linearized Attention (LA)} \label{linear_attention_section_appendix}
\begin{figure}
\centering
 \begin{subfigure}[t]{0.5\textwidth}
 \centering
    \includegraphics[width=1.0\columnwidth]{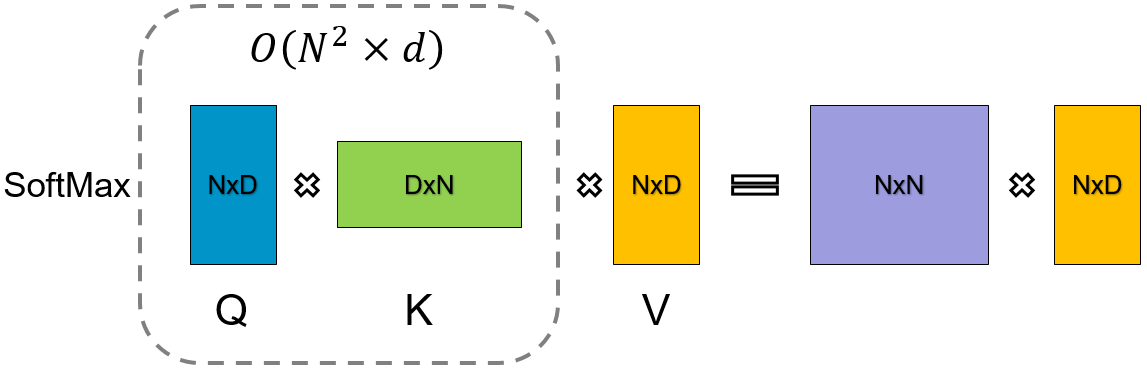}
    \caption{Softmax attention}
    \label{fig:softmax_attention_appendix}
 \end{subfigure}
 \hfill
 \begin{subfigure}[t]{0.38\textwidth}
 \centering
    \includegraphics[width=1.00\columnwidth]{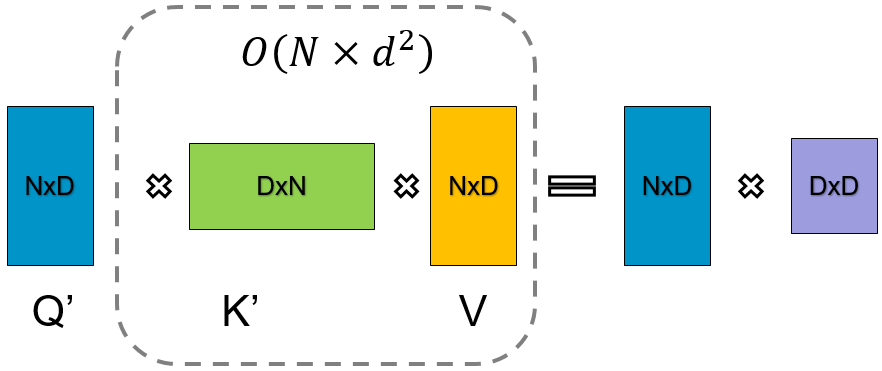}
    \caption{Linear attention}
    \label{fig:linear_attention_appendix}
 \end{subfigure}
 \caption{A block diagram of the computational complexity for the Softmax Attention and Linearized Attention.}
\end{figure}

Let $\mb{Q},\mb{K},\mb{V} \in \mathbb{R}^{N \times d}$ denote the queries, keys, and values of the attention mechanism. Computing softmax-attention with Equation \eqref{eq:softmax_attention_appendix} requires calculating the quadratic matrix $\mb{Q}\mb{K}^\top=\left[\mb{q}_i^\top \mb{k}j\right]_{N \times N}$, which has a complexity of $\mathcal{O}(N^2)$ with respect to the sequence length $N$ as illustrated in Figure \ref{fig:softmax_attention_appendix}.

Alternatively, if we can decompose $\rm{softmax}\left(\mb{Q}\mb{K}^\top\right)$ into $\mb{Q}'\mb{K}'^\top$, we can use the associativity property of matrix multiplication to compute $\mb{Q}'\left(\mb{K}'^\top\mb{V}\right)$ from right to left. Such a decomposed computation has a linear complexity in the sequence length $N$, as illustrated in Figure \ref{fig:linear_attention_appendix}. To obtain this decomposition, the un-normalized matrix $e^{\mb{Q}\mb{K}^\top}$ is replaced with matrix $\Phi(\mb{Q})\Phi(\mb{K})^\top$, where $\Phi$ is a feature map that is applied row-wise, i.e., $\mb{Q}'=\Phi(\mb{Q})$ and $\mb{K}'=\Phi(\mb{K})$. This form of linear attention can be expressed as follows:
\begin{equation}
    \text{Attn}_{\text{lin}}(\mb{q}_i, \{\mb{k}_1, \dots, \mb{k}_N\}, \{\mb{v}_1, \dots, \mb{v}_N\}) = \frac{\sum_{j=1}^N \Phi(\mb{q}_i)^\top \Phi(\mb{k}_j)}{\sum_{l=1}^N \Phi(\mb{q}_i)^\top\Phi(\mb{k}_l)}\mb{v}^\top_j = \frac{\Phi(\mb{q}_i)^\top\sum_{j=1}^N \Phi(\mb{k}_j) \mb{v}^\top_j}{\Phi(\mb{q}_i)^\top\sum_{l=1}^N \Phi(\mb{k}_l)} 
    \label{eq:linearized_attn_appendix}
\end{equation}

Due to aggregation over $N$ elements in both the numerator and denominator, the memory complexity of this linearized attention mechanism is linear with respect to the sequence length $N$.

\subsection{Kernel view of the self-attention}
The SA mechanism can be viewed as a Nadaraya-Watson kernel regression \citep{nadaraya1964estimating} as shown by \citet{Transformers-RKBS}. Specifically, the kernel density estimation (KDE) of some unknown function with joint distribution $p(\mb{k},\mb{v})$ and density $p(\mb{k})$ with a kernel $\kappa$ \citep{Robustify-Han}. Therefore, the kernel function can be used to express the self-attention mechanism as follows:

\begin{equation}
    \text{Attn}(\mb{q}_i, \{\mb{k}_1, \dots, \mb{k}_N\}, \{\mb{v}_1, \dots, \mb{v}_N\}) = \frac{\sum_{j=1}^N \kappa(\mb{q}_i, \mb{k}_j) \mb{v}^\top_j}{\sum_{l=1}^N \kappa (\mb{q}_i, \mb{k}_l)} .
    \label{eq:kde_attn_appendix}
\end{equation}

he KDE form of \Cref{eq:kde_attn_appendix} is a generalization of softmax attention in \Cref{eq:softmax_attention_appendix} where we use a softmax kernel 
\begin{align}
    \kappa_{\rm sm}(\mb{q}_i,\mb{k}_j)=e^{\frac{\innerproduct{\mb{q}_i}{\mb{k}_j}}{\sqrt{d}}}
\label{eq:softmax_attn_kernel_appendix}
\end{align}
Interestingly, the kernel view of self-attention reveals that the functionality of the attention mechanism remains intact even when the kernel is modified. However, the performance of the attention mechanism can vary depending on the choice of the kernel function, as demonstrated by \citet{Yao-Transformer-Dissection}. It emphasizes the significance of considering different kernels and their impact on the performance of linearized attention. Additionally, the kernel view provides a valuable perspective for tackling the linearization problem within the framework of the kernel method.

According to Mercer's theorem \citep{mercer1909functions}, any positive-definite kernel (Mercer kernel) can be represented as an inner product of symmetric features. Let $\mathcal{X} \in \mathbb{R}^{N \times d}$, $\mb{x}_1, ..., \mb{x}_N \in \mathcal{X}$ with kernel function $\kappa: \mathcal{X}\times\mathcal{X} \to \mathbb{R}$, then
\begin{equation}
    \kappa(\mb{x}_i, \mb{x}_j) = \innerproduct{\Phi(\mb{x}_i)}{\Phi(\mb{x}_j)}_{\mathcal{F_H}}
\end{equation}
where $\Phi:\mathcal{X} \to \mathcal{F_H}$ is a function mapping the inputs to a Hilbert space of feature functions $\mathcal{F_H}$. However, the dimensionality of $\mathcal{F_H}$ can be large or infinite, making explicit computation of the features infeasible. This is where the kernel trick comes in: the kernel function can be computed without explicitly computing the features. However, storing the attention matrix still requires $\mathcal{O}(N^2)$ memory complexity. To address this issue, we can design a kernel function such that the dimensionality of $\mathcal{F_H}$ is much smaller than $N$, which allows for $\mathcal{O}(N)$ memory complexity. For example, consider $\Phi_\mathcal{Q}: \mathcal{Q} \to \mathbb{R}^d$ and $\Phi_\mathcal{K}: \mathcal{K} \to \mathbb{R}^d$ for some $d \ll N$, the attention can be computed using the following linear kernel function:
\begin{align}
    \kappa(\mb{q}_i, \mb{k}_j)=\innerproduct{\Phi_\mathcal{Q}(\mb{q}_i)}{\Phi_\mathcal{K}(\mb{k}_j)}
    \label{eq:linear_kernel_appendix}
\end{align}
Consequently, using the associativity property described in section \ref{linear_attention_section_appendix} can be used to compute this kernel function with $\mathcal{O}(N)$ memory complexity.

\subsection{Analysis of the entropy, variance and spectral gap of the Softmax Attention marix}\label{sec:variance_of_the_softmax}

We start our analysis by providing additional definitions which extend those in \Cref{sec:analysis} and proving a couple of lemmas. Denote by $\mb{a} \in \mathbb{R}^N$ a single row of the attention scores matrix $\mb{A}$. By denoting a corresponding row of the normalized attention scores matrix $\mb{\Tilde{A}}$ by $\mb{x}$, such that $\tau \mb{a} = \mb{x}$, we can write $\mb{p} \in \mathbb{R}^N$ a single row of the softmax attention matrix from \cref{eq:softmax_attention_matrix} as:
\begin{align} 
    \mb{p} = \text{softmax}(\mb{x}, \tau) = \frac{e^\frac{\mb{x}}{\tau}}{\sum_{j=1}^N e^\frac{x_j}{\tau}}
    \label{eq:softmax_with_temperature_apendix}
\end{align}
Where $\tau$ is the temperature of SA from \cref{eq:temperature_of_softmax_attention}.
Similarly to \cref{eq:softmax_attention_entropy}, we define the entropy of a single row $\mb{p}$ as:
\begin{align}
    H(\mb{p}) = -\sum_{i=1}^N p_i \log_2(p_i)
    \label{eq:entropy_single_row_appndx}
\end{align}

Additionally, we define a variance of the single row $\mb{p}$ of the SA matrix as:
\begin{align}
    \sigma_\mb{p}^2 = \frac{1}{N}\sum_{i=1}^N \left(p_i - \frac{1}{N}\sum_{j=1}^Np_j\right)^2 = \frac{1}{N}\sum_{i=1}^N \left(p_i - \frac{1}{N}\right)^2
\end{align}

Denote $\mu = \sum_{i=1}^N p_i x_i$

\begin{lemma}
\label{lemma:simple_mean_zero}
The following holds \(\sum_{i=1}^N p_i(x_i - \mu)=0\)
\begin{proof}
\begin{align*}
    \sum_{i=1}^N p_i(x_i - \mu) = \sum_{i=1}^N p_ix_i - \mu \underbrace{\sum_{i=1}^N p_i}_{1} = \mu - \mu = 0
\end{align*}
\end{proof}
\end{lemma}

\begin{lemma}
\label{lemma:quadratic_prob_exp}
Let $\mb{p}$ as in \cref{eq:softmax_with_temperature_apendix}, then the following holds.
\begin{align}
    \sum_{i=1}^N p_i^2 (x_i - \mu) \ge 0
    \label{eq:lemma_quad_prob_positive}
\end{align}

\begin{proof}
By denoting \(\delta_i = x_i - \mu\) and substituting it into \Cref{eq:lemma_quad_prob_positive} together with \cref{eq:softmax_with_temperature_apendix}, we obtain
\begin{align*}
    \sum_{i=1}^N p_i^2 (x_i - \mu) = \frac{1}{\left(\sum_{j=1}^N e^\frac{x_j}{\tau}\right)^2} \sum_{i=1}^N e^\frac{2x_i}{\tau} \delta_i = \frac{1}{\left(\sum_{j=1}^N e^\frac{\delta_j}{\tau}\right)^2} \sum_{i=1}^N e^\frac{2\delta_i}{\tau} \delta_i
\end{align*}
Note that we can replace $x_i$ by $\delta_i$ due to the translation invariance of the softmax. Thus, it remains to show that:
\begin{align*}
    \sum_{i=1}^N e^\frac{2\delta_i}{\tau}\delta_i \ge 0
\end{align*}
By expressing the equation from \Cref{lemma:simple_mean_zero} in terms of $\delta_i$ we get
\begin{align*}
    \sum_{i=1}^N p_i(x_i - \mu) = \frac{1}{\sum_{j=1}^N e^\frac{x_j}{\tau}} \sum_{i=1}^N e^\frac{x_i}{\tau} \delta_i = \frac{1}{\sum_{j=1}^N e^\frac{\delta_j}{\tau}} \sum_{i=1}^N e^\frac{\delta_i}{\tau} \delta_i = 0
\end{align*}
Thus,
\begin{align}
    \sum_{i=1}^N e^\frac{\delta_i}{\tau}\delta_i = 0
    \label{eq:exp_prob_sum}
\end{align}
Split the sum in \cref{eq:exp_prob_sum} into two parts, first sum over the elements $\delta_i \ge 0$ and second over $\delta_i < 0$, as following:
\begin{align*}
    0 &= \sum_{i=1}^N e^\frac{\delta_i}{\tau} \delta_i = \sum_{\delta_i \ge 0} e^\frac{\delta_i}{\tau} \delta_i + \sum_{\delta_i < 0} e^\frac{\delta_i}{\tau} \delta_i \le \\
    &\le \sum_{\delta_i \ge 0} \underbrace{e^\frac{\delta_i}{\tau}}_{\ge 1} e^\frac{\delta_i}{\tau} \delta_i + \sum_{\delta_i < 0} \underbrace{e^\frac{\delta_i}{\tau}}_{< 1} e^\frac{\delta_i}{\tau} \delta_i = \\
    &= \sum_{i=1}^N e^\frac{\delta_i}{\tau} e^\frac{\delta_i}{\tau} \delta_i = \sum_{i=1}^N e^\frac{2\delta_i}{\tau} \delta_i = \sum_{i=1}^N p_i^2 (x_i - \mu)
\end{align*}
\end{proof}
\end{lemma}

\begin{theorem}
The variance $\sigma_\mb{p}^2$ is monotonically decreasing with temperature $\tau$.
\begin{proof}
By taking the derivative of the $\sigma_\mb{x}^2$ with respect to the $\tau$ we get.

\begin{align*}
    \derv{\sigma_\mb{p}^2}{\tau} &= 2 \frac{1}{N}\sum_{i=1}^N (p_i - \frac{1}{N}) \derv{p_i}{\tau} = -2 \frac{1}{N}\sum_{i=1}^N (p_i - \frac{1}{N}) p_i(x_i - \sum_j^N x_j p_j) \frac{1}{\tau^2} = \\
    &= - \frac{1}{N} \frac{2}{\tau^2} \sum_{i=1}^N \left(p_i^2 x_i -\frac{1}{N} p_i x_i - p_i^2 \mu + \frac{1}{N} p_i \mu \right) = \\
    &= - \frac{2}{\tau^2} \frac{1}{N}\left(\sum_{i=1}^N p_i^2 x_i - \mu \sum_{i=1}^N p_i^2 - \frac{1}{N} \underbrace{ \sum_{i=1}^N p_i(x_i - \mu) }_{0} \right) = \\
    &= - \frac{2}{\tau^2} \frac{1}{N} \underbrace{\sum_{i=1}^N p_i^2 (x_i - \mu)}_{\ge 0} < 0
\end{align*}
Note that \(\sum_{i=1}^N p_i^2 (x_i - \mu) \ge 0\) as follows from the \Cref{lemma:quadratic_prob_exp}.
\end{proof}
\label{theorem:var_temp_proof_appendix}
\end{theorem}

\begin{lemma}
\label{lemma:prob_var_expression}
The following holds
\begin{equation*}
    \sum_{i=1}^N p_i x_i^2 - \mu^2 = \sum_{i=1}^N p_i (x_i - \mu)^2
\end{equation*}
\begin{proof}
\begin{align*}
\sum_{i=1}^N p_i x_i^2 - \mu^2 &= \sum_{i=1}^N p_i (x_i - \mu + \mu)^2 - \mu^2 = \\
&= \sum_{i=1}^N p_i (x_i - \mu)^2 + 2\sum_{i=1}^N p_i(x_i - \mu)\mu + \mu^2 - \mu^2 = \\
&= \sum_{i=1}^N p_i (x_i - \mu)^2 + 2 \mu \underbrace{\sum_{i=1}^N p_i(x_i - \mu) }_{0} = \sum_{i=1}^N p_i (x_i - \mu)^2
\end{align*}
\end{proof}
\end{lemma}

\subsubsection{Proof of \texorpdfstring{\Cref{theorem:entropy_temperature_monotonicity}}{}}
\label{proof:entropy_temperature_monotonicity}

\begin{proof}
To show that the entropy in \cref{eq:softmax_attention_entropy} is monotonically increasing with temperature, we first show that the entropy of the single row \cref{eq:entropy_single_row_appndx} of the SA matrix is monotonically increasing. To that end, we take a derivative $\derv{H}{\tau}$ of the entropy with respect to the temperature and show that is always positive.

Denote $S = \sum_{j=1}^N e^\frac{x_j}{\tau}$

The derivative of the single entry of $\mb{p}$ with respect to the temperature is given by:
\begin{align*}
\derv{p_i}{\tau} &= \frac{(-\frac{x_i}{\tau^2}) e^\frac{x_i}{\tau} S - e^\frac{x_i}{\tau} \sum_{j=1}^N e^\frac{x_j}{\tau} (-\frac{x_j}{\tau^2}) }{S^2} = \\
 &= -\frac{x_i}{\tau^2} p_i - p_i \sum_{j=1}^N p_j (-\frac{x_j}{\tau^2}) = \\
 &= -\frac{1}{\tau^2}p_i(x_i - \sum_j^N x_j p_j) = \\
 &= -\frac{1}{\tau^2}p_i(x_i - \mu) 
\end{align*}

The derivative of the entropy from \cref{eq:entropy_single_row_appndx} with respect to the temperature:

\begin{align*}
    \derv{H}{\tau} &= - \sum_{i=1}^N \left(\log_2(p_i) + \frac{1}{\ln2}\right) \derv{p_i}{\tau} = \sum_{i=1}^N \left(\log_2(p_i) + \frac{1}{\ln2}\right) \frac{1}{\tau^2}p_i(x_i - \mu) = \\
    &= \begin{aligned}[t]
    \frac{1}{\tau^2} \Biggl( \sum_{i=1}^N p_i x_i \log_2(p_i) - \sum_{i=1}^N p_i \log_2(p_i) \mu + \frac{1}{\ln2} \sum_{i=1}^N p_i x_i - \sum_{i=1}^N p_i \frac{1}{\ln2} \mu \Biggr)
    \end{aligned} = \\
    &= \frac{1}{\tau^2} \left( \sum_{i=1}^N p_i x_i \log_2(p_i) - \mu \sum_{i=1}^N p_i \log_2(p_i) \right) = \\
    &= \frac{1}{\tau^2}  \sum_{i=1}^N p_i \log_2(p_i) (x_i - \mu) = \\
    &= \frac{1}{\tau^2}  \sum_{i=1}^N p_i \left( \frac{x_i}{\tau} - \log_2 S \right) (x_i - \mu) = \\
    &= \frac{1}{\tau^2} \left( \frac{1}{\tau}\sum_{i=1}^N p_i x_i^2 - \log_2 S \sum_{i=1}^N p_i x_i + \log_2 S \mu - \frac{1}{\tau} \mu \sum_{i=1}^N p_i x_i  \right) = \\
    &= \frac{1}{\tau^3} \left(\sum_{i=1}^N p_i x_i^2 - \mu^2  \right) \underbrace{=}_{\ast} \frac{1}{\tau^3} \sum_{i=1}^N p_i (x_i - \mu)^2 > 0
\end{align*}
Where, $\ast$ follows from \Cref{lemma:prob_var_expression}.

Finally, since the entropy in \cref{eq:softmax_attention_matrix} is the average entropy of the rows, the \Cref{theorem:entropy_temperature_monotonicity} follows.
\begin{align}
    H(\mb{P}^{\text{(SM)}}) = \frac{1}{N}\sum_{i=1}^N H(\mb{p}_i)
\end{align}

\end{proof}

\subsubsection{Proof of \texorpdfstring{\Cref{theorem:lambda2_var_max}}{}}
\label{proof:lambda2_var_max}

The proof is based on the Principal Component Analysis (PCA) \citep{Jolliffe2011} of the stochastic matrix $\mb{P}$.
\begin{proof}

To start our proof, we first need to center matrix $\mb{P}$ in \cref{eq:softmax_attention_matrix} by deflating the first eigenvalue. Let $\pmb{\mu} = \frac{1}{N} \mb{P}^\top \mb{1} = \frac{1}{N} \sum_{i=1}^N \mb{P}_{ij}$ the vector mean of $\mb{P}$. Note that $\mb{v}_1 = \mb{1}$ is the eigenvector of $\mb{P}$ corresponding to the first eigenvalue $\lambda_1=1$ for which the following holds:
\begin{align}
    \pmb{\mu}^\top \mb{v}_1 = \frac{1}{N} \mb{1}^\top \mb{P} \mathbf{1} = \frac{1}{N} \sum_{i=1}^N \sum_{j=1}^N \mb{P}_{ij} = 1
\end{align}
Thus, according to the Wielandt deflation theorem the eigenvalues of the matrix $\overline{\mb{P}} = \mb{P} - \lambda_1 \mb{v}_1 \pmb{\mu}^\top$ are $0, \lambda_2, \dots, \lambda_N$. 

Furthermore, it is important to note that the matrix $\overline{\mb{P}}$ is centered in both rows and columns as follows:
\begin{align*}
    \sum_i^N \overline{\mb{P}}_{ij} = \mb{1}^\top \overline{\mb{P}} = \mb{1}^\top \mb{P} - \lambda_1 \mb{1}^\top \mb{v}_1\pmb{\mu}^\top = \mb{1}^\top \mb{P} - N \mb{\mu}^\top = \mb{1}^\top \mb{P} - \mb{1}^\top \mb{P} = 0
\end{align*}
\begin{align*}
    \sum_j^N \overline{\mb{P}}_{ij} = \overline{\mb{P}}\mb{1} = \mb{P}\mb{1} - \lambda_1 \mb{v}_1 \pmb{\mu}^\top \mb{1} = \mb{1} - \mb{v}_1 \frac{1}{N} \mb{1}^\top\mb{P}  \mb{1} = \mb{1} - \mb{1} = 0
\end{align*}

Therefore, the empirical covariance matrix of $\mb{P}$ can be expressed as:

\begin{align*}
    \mb{\Sigma}_{\mb{P}} = \left(\mb{P} - \lambda_1 \mb{v}_1 \pmb{\mu}^\top\right)^\top\left(\mb{P} - \lambda_1 \mb{v}_1 \pmb{\mu}^\top\right)  = \overline{\mb{P}}^\top\overline{\mb{P}}
\end{align*}

Denote by $\mb{\Bar{v}}_1, \dots, \mb{\Bar{v}}_N$ the eigenvectors of $\overline{\mb{P}}$, where $\mb{\Bar{v}}_1$ corresponds to the eigenvalue $\lambda_1=0$.

The variance in $\mb{P}$ along the direction of the principal component $\mb{\Bar{v}}_i$ can be expressed as:

\begin{align*}
    \sigma^2_{\mb{\Bar{v}}_i} &= \frac{\mb{\Bar{v}}_i^\ast \mb{\Sigma}_{\mb{P}} \mb{\Bar{v}}_i}{\mb{\Bar{v}}_i^\ast \mb{\Bar{v}}_i} = \mb{\Bar{v}}_i^\ast \overline{\mb{P}}^\top\overline{\mb{P}} \mb{\Bar{v}}_i = \norm{\overline{\mb{P}} \mb{\Bar{v}}_i}^2 = \lambda_i^2 \norm{\mb{\Bar{v}}_i}^2 = \lambda_i^2
\end{align*}
Therefore, the variance is maximized when $\mb{\Bar{v}}_i = \mb{\Bar{v}}_2$, and $\lambda_2^2 = \sigma^2_{\mb{\Bar{v}}_2}$ represents the amount of variance in the direction specified by the largest principal component of $\overline{\mb{P}}$.
\end{proof}

Thus, $\lambda_2$ indicates the level of variability of matrix $\mb{P}$ along the direction specified by the major principal component of $\overline{\mb{P}}$. Moreover, according to \Cref{theorem:var_temp_proof_appendix}, if the rank of matrix $\mb{P}$ remains unchanged, the variability should decrease as the temperature increases. However, if the matrix $\mb{P}$ is biased towards a particular column, the variability within the columns decreases, resulting in a smaller value of $\lambda_2$. Consequently, we can infer that the spectral gap, defined as \(\gamma = 1 - |\lambda_2| = 1 - \sigma_{\mb{\Bar{v}}_2} \le 1\), increases with the temperature only when the stochastic matrix is unbiased.

\subsection{Proof of \texorpdfstring{\Cref{proposition:softmax_distribution}}{}}
\label{section:softmax_distribution_appendix}

\begin{proof}
Generally, the product of two independent Gaussian variables has a density in the form of a modified Bessel function of the second kind \citep{weisstein2003normal}. When the vector dimensions are sufficiently large, the Central Limit Theorem implies that the distribution of the dot product between $\mb{q}$ and $\mb{k}$ can be approximated by a Gaussian distribution with zero mean and variance $\sigma^2$. As mentioned in \cref{sec:analysis}, the variance of $\mb{q}^\top \mb{k}$ can be expressed as \(\sigma^2 = \sigma^2_q\sigma^2_k + C_{\rm cross}\), where \(C_{\rm cross} = \text{Cov}(\mb{q}^2, \mb{k}^2) - \text{Cov}(\mb{q}, \mb{k})^2\) is the cross-covariance of the squared queries and keys \citep{variance_of_products}. 

Therefore, due to the exponent in \cref{eq:softmax_attention_matrix}, the numerator is approximately a log-normal variable with zero mean and variance $\sigma^2$. To address the denominator, we must consider a sum of log-normal variables. Fortunately, \citet{Fenton1960TheSO} theorem states that for moderate values of $\sigma^2$, the sum of zero-mean i.i.d. log-normal variables can be approximated by a log-normal variable with variance $\sigma_\Sigma^2$ and mean $\mu_\Sigma$ where:
\begin{align*}
    \sigma_\Sigma^2 = \ln\left( \frac{1}{N} \left(e^{\sigma^2} - 1\right) +1 \right); \quad \mu_\Sigma = \ln N + (\sigma^2 - \sigma_\Sigma^2)/2
\end{align*}

For large $N$ and moderate values of $\sigma^2$, we have $\sigma_\Sigma^2 \ll \sigma^2$, and we can omit the $\sigma_\Sigma^2$ term for simplicity. Finally, since the ratio of log-normal variables remains log-normal with mean $-\mu_\Sigma$ and variance $\sigma^2$, \Cref{proposition:softmax_distribution} follows.
\end{proof}

\begin{figure}
\centering
 \begin{subfigure}[t]{0.49\textwidth}
 \centering
    \includegraphics[width=1.0\columnwidth]{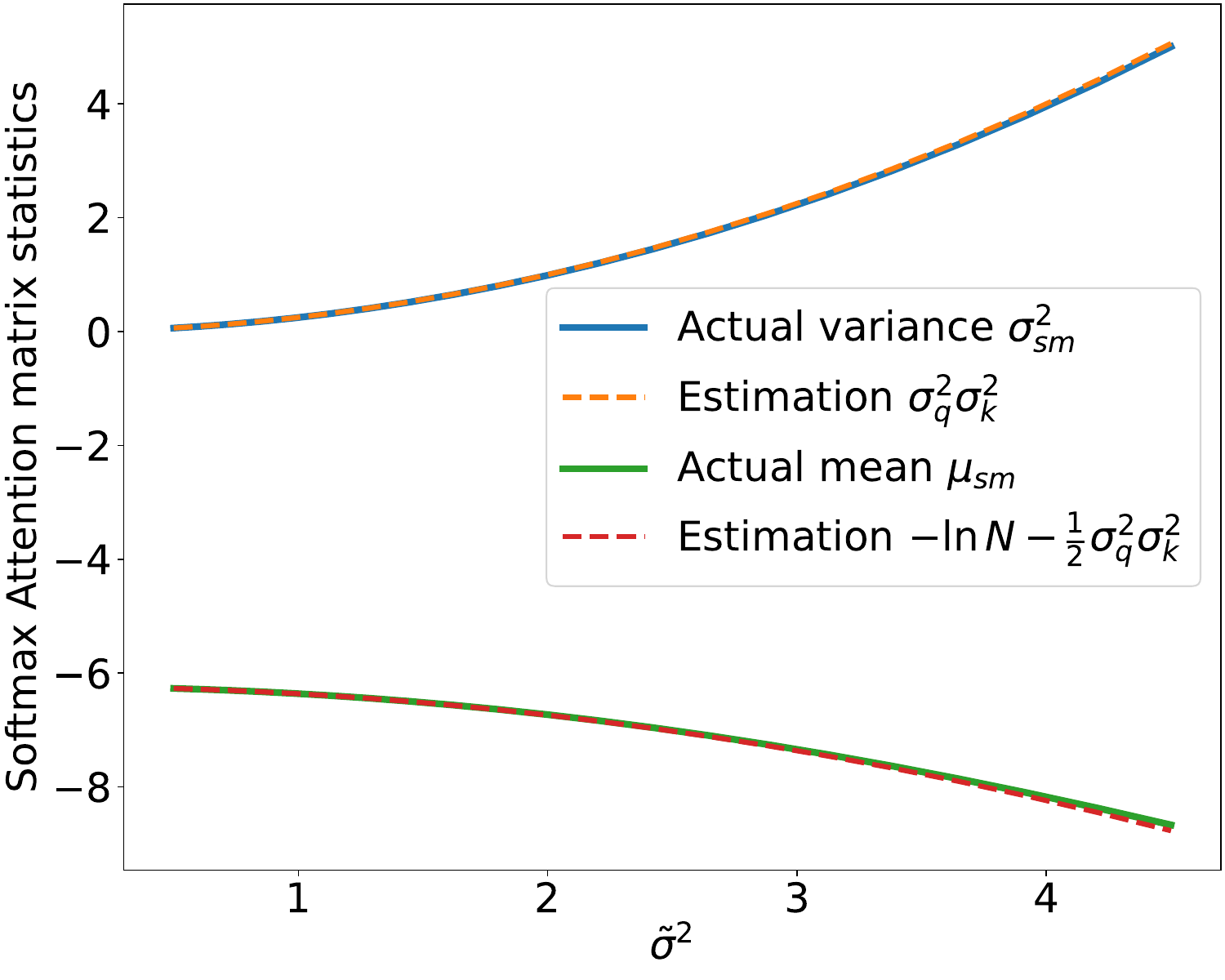}
    \caption{}
\label{fig:softmax_statistics}
 \end{subfigure}
 \hfill
 \begin{subfigure}[t]{0.49\textwidth}
 \centering
    \includegraphics[width=1.00\columnwidth]{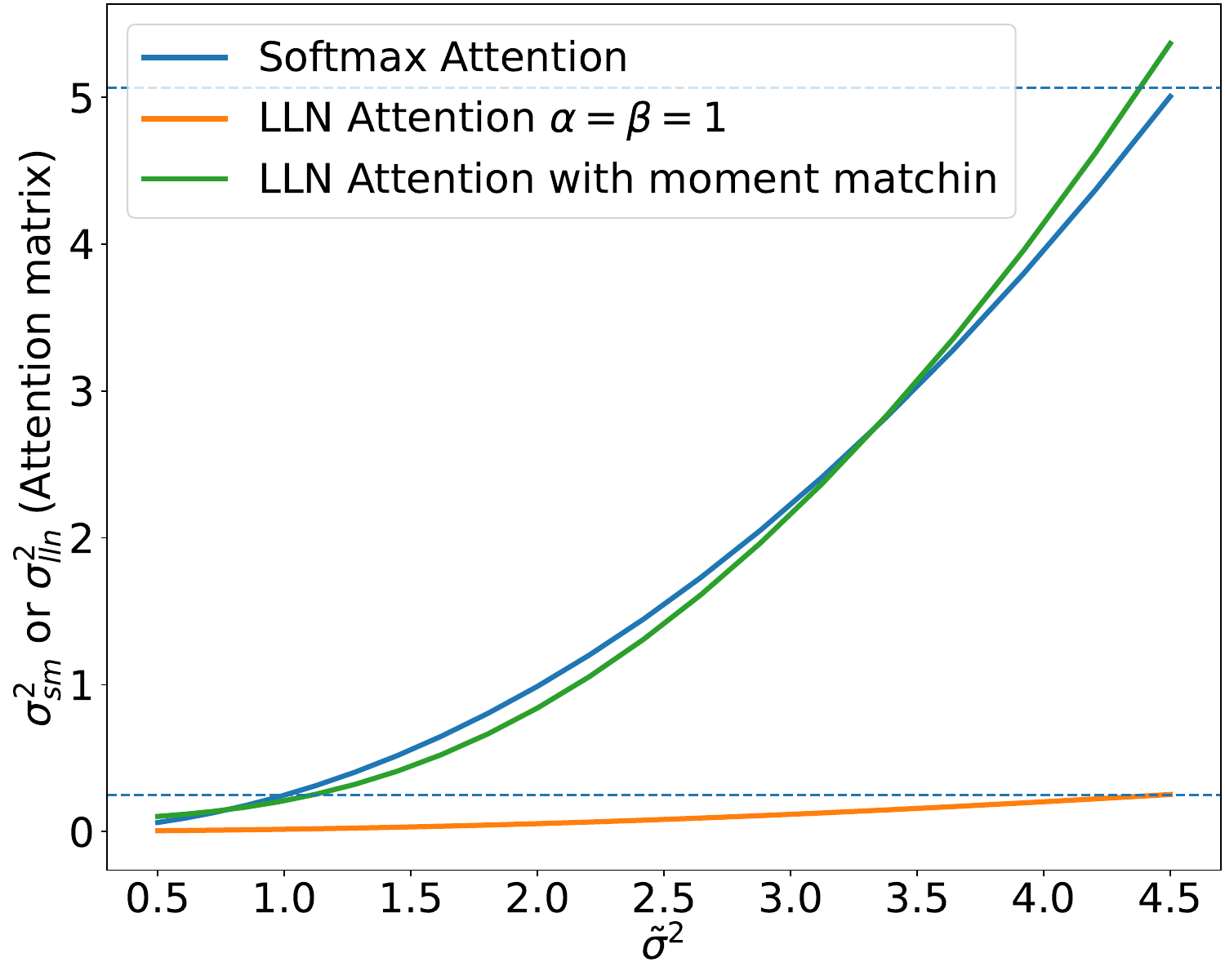}
    \caption{}
 \label{fig:lln_moment_matching}
 \end{subfigure}
 \caption{(a) The variance and mean of the SA matrix with respect to the input variance. Measurements perfectly match theoretical estimation. (b) The variance of the SA and LLN Attention before and after performing the moment matching procedure.}
\end{figure}

To empirically validate the assumption made in \Cref{proposition:softmax_distribution}, we measure the variance and mean of the SA and compare them to the values predicted by the theory. The results are presented in \Cref{fig:softmax_statistics}, which shows that the measured statistics closely match the theoretical predictions.
  

\subsection{Proof of \texorpdfstring{\Cref{proposition:lln_distribution}}{}}
\label{proposition:lln_distribution_proof}
\begin{proof}
To demonstrate that the LLN attention matrix approximately follows a log-normal distribution, we examine a single entry of the attention matrix, which is a dot product of two vectors $\mb{q}$ and $\mb{k}$. The nominator $\innerproduct{e^{\alpha \mb{q}}}{e^{\beta \mb{k}}}=\sum_i^d e^{\alpha q_i + \beta k_i}$ represents a sum of $d$ log-normal variables with zero mean and variance $\Tilde{\sigma}^2$, where
\begin{align}
    \Tilde{\sigma}^2 = \alpha^2\sigma_q^2+\beta^2\sigma_k^2
\end{align}
Similarly, the denominator $\sum_{j=1}^N \innerproduct{e^{\alpha \mb{q}}}{e^{\beta \mb{k}_j}}$ of the LLN Attention matrix, which is also a sum of log-normal variables. According to \citep{Fenton1960TheSO}, for moderate values of $\Tilde{\sigma}^2$, the distribution of the sum of log-normal variables can be approximated by another log-normal distribution at the right tail. Since the ratio of the log-normal variables is also log-normal, we can approximate the distribution of the LLN Attention matrix by a log-normal distribution.

To determine the relationship between the variance of the LLN Attention matrix $\sigma_{LLN}^2$ and variances of queries and keys $\sigma_q^2$, $\sigma_k^2$, we need to estimate the variance of a sum of log-normal variables. Following the approach of \citep{Romeo_2003_lognormal_sum}, we will divide our analysis into three cases: narrow $\Tilde{\sigma}^2 \ll 1$, moderate $\Tilde{\sigma}^2 \lesssim 1$ and broad $\Tilde{\sigma}^2 > 1$.

Denote variance of the sum in nominator by $\sigma_{nom}^2$ and variance of the denominator by $\sigma_{den}^2$. 

\subsubsection*{Narrow case}
If $0 < \sigma_q^2, \sigma_k^2 \ll 1$ then the values are small such that even close to zero, thus we can approximate $e^{\alpha q_i} \approx 1 + \alpha q_i$ and $e^{\beta k_i} \approx 1 + \beta k_i$, thus:

\begin{equation}
    \sigma_{nom}^2 \approx d (\alpha^2 \sigma^2_q + \beta^2 \sigma^2_k) = d \Tilde{\sigma}^2; \quad \sigma_{den}^2 \approx N (\alpha^2 \sigma^2_q + \beta^2 \sigma^2_k) = N \Tilde{\sigma}^2
    \label{eq:var_lln_narrow}
\end{equation}

\subsubsection*{Moderate case}
When $\Tilde{\sigma}^2$ is relatively small (i.e., $\lesssim 1$), we can use \citep{Fenton1960TheSO} method. The approximation is given by:

\begin{equation}
    \sigma_{nom}^2 \approx \ln \left[ \frac{\left(e^{\alpha^2 \sigma^2_q + \beta^2 \sigma^2_k} - 1\right)}{d} + 1\right] = \ln \left[ \frac{\left(e^{\Tilde{\sigma}^2} - 1\right)}{d} + 1\right]
    \label{eq:var_lln_moderate}
\end{equation}
Similarly, for the denominator:
\begin{equation}
    \sigma_{den}^2 \approx \ln \left[ \frac{\left(e^{\Tilde{\sigma}^2} - 1\right)}{N} + 1\right]
\end{equation}

In \cref{fig:lognormal_sum_modelrate}, we demonstrate the empirical evaluation of the correctness of Fenton approximation when $\sigma^2 \in [0.2,1.2]$.

\begin{figure}
\centering
 \begin{subfigure}[t]{0.49\textwidth}
 \centering
    \includegraphics[width=1.0\columnwidth]{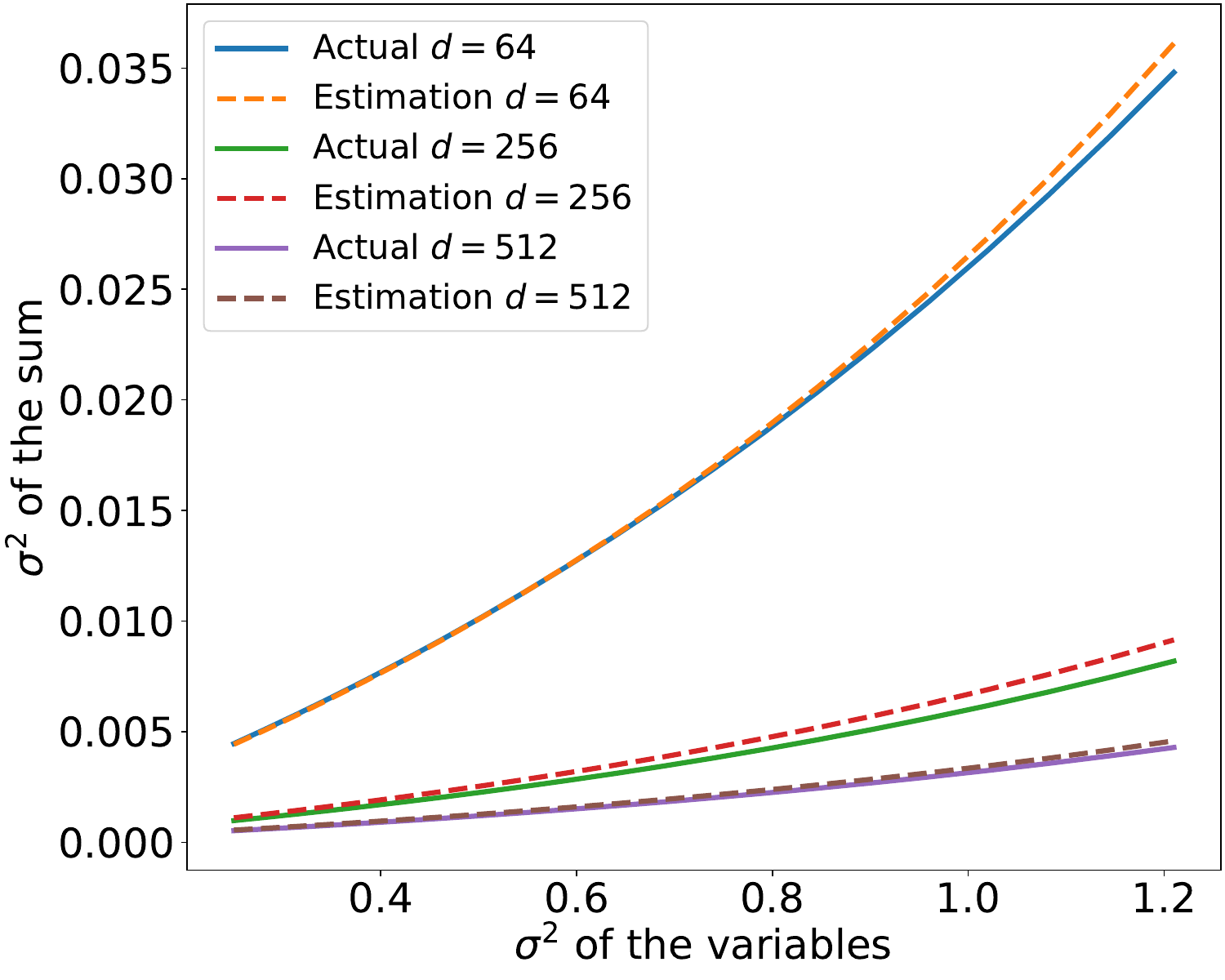}
    \caption{}
\label{fig:lognormal_sum_modelrate}
 \end{subfigure}
 \hfill
 \begin{subfigure}[t]{0.49\textwidth}
 \centering
    \includegraphics[width=1.00\columnwidth]{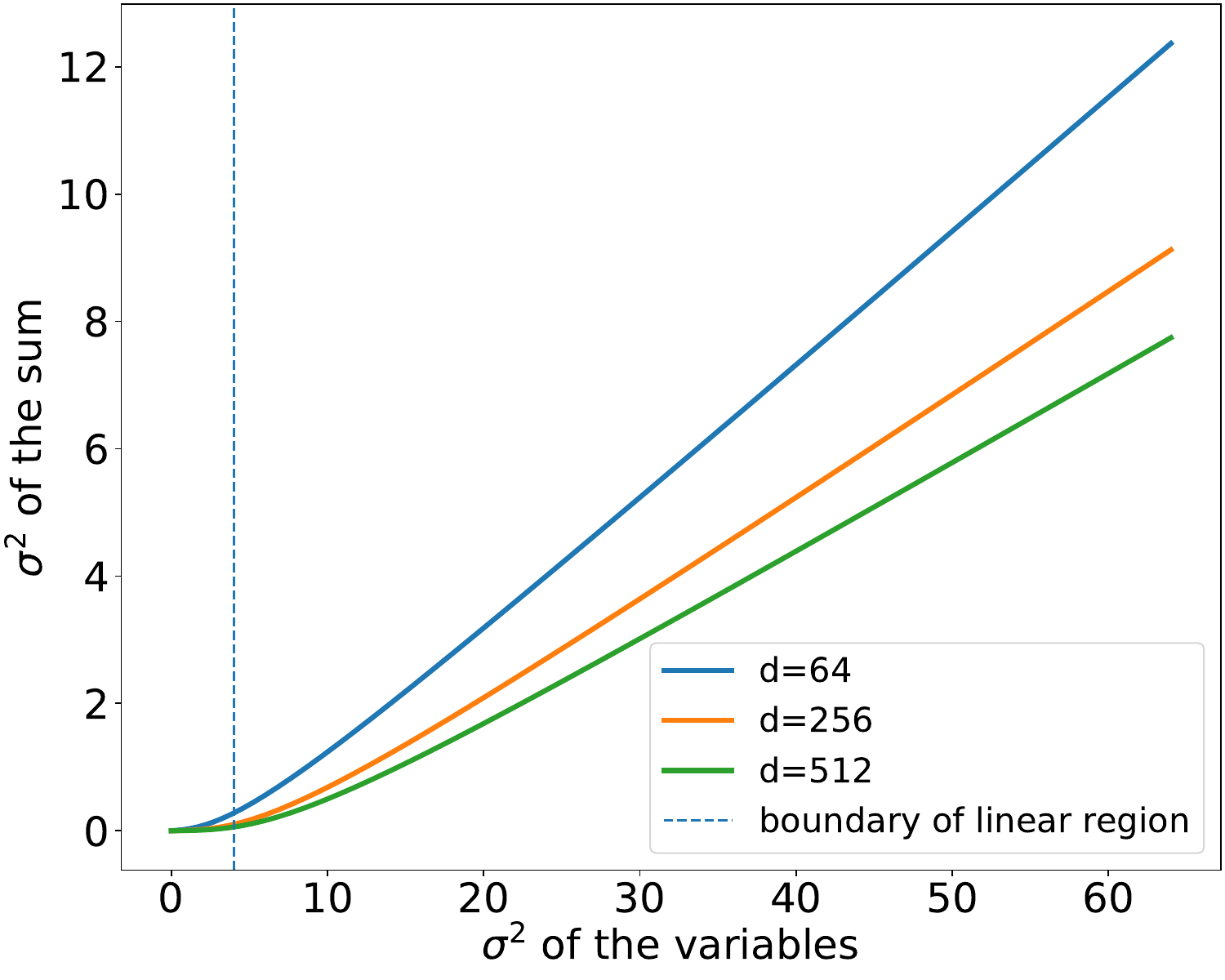}
    \caption{}
 \label{fig:lognormal_sum_broad}
 \end{subfigure}
 \caption{The variance of the sum of $d$ log-normally distributed inputs, each with a variance of $\sigma^2$ and a zero mean, is effectively estimated by the \citep{Fenton1960TheSO} method. (a) In moderate scenario where $\sigma^2$ lies within the range of $[0.2, 1.2]$, theoretical estimations (depicted as dashed lines) aligned with empirical results. 
 (b) In the broad case, for sufficiently large $\sigma^2$ the variance of the log-normal sum grows linearly with the variance of the input variables.}
\end{figure}

\subsubsection*{Broad case}
In the broad case, when $\Tilde{\sigma}^2$ is large (i.e., $\Tilde{\sigma}^2 \gg 1$), it is not possible to find a closed-form approximation for the log-normal sum. However, we can observe that the sum of exponents is dominated by the largest term, which corresponds to the maximum value. This maximum value grows linearly with the spread of queries and keys under the Gaussian assumption. Consequently, according to \citep{Romeo_2003_lognormal_sum}, the resulting variance $\Tilde{\sigma}^2$ is also linearly proportional to $\Tilde{\sigma}^2$ with some constants $a_1, a_2$ and $b_1, b_2$:
\begin{equation}
    \sigma^2_{nom} \approx a_1 (\alpha^2\sigma^2_q + \beta^2 \sigma^2_k) + b_1 = a_1 \Tilde{\sigma}^2 + b_1
\end{equation}
\begin{equation}
    \sigma^2_{den} \approx a_2 (\alpha^2\sigma^2_q + \beta^2 \sigma^2_k) + b_2 = a_2 \Tilde{\sigma}^2 + b_2
\end{equation}

We empirically evaluated the linear dependency assumption of the sum of log-normally distributed inputs and showed its validity in \cref{fig:lognormal_sum_broad}.

Finally, the variance of the LLN Attention matrix is a sum of the nominator and denominator variances, i.e.:

\begin{equation}
    \sigma_\text{lln}^2 = \sigma^2_{nom} + \sigma^2_{den}
\end{equation}
By denoting $a = a_1 + a_2$ and $b = b_1 + b_2$ for a broad case we get:
\begin{equation}
    \sigma_\text{lln}^2 = a \Tilde{\sigma}^2 + b = a(\alpha^2 \sigma^2_q + \beta^2 \sigma^2_k) + b
\end{equation}

\end{proof}

\subsection{Moment matching}\label{sec:moment_matching}
As shown in \Cref{fig:lln_moment_matching}, we are interested in handling a broad range of $\Tilde{\sigma}^2$ values, particularly those in $[1,4]$. Using \Cref{proposition:lln_distribution}, we can find the constants $a$ and $b$ that satisfy the requirement $\sigma_{lln}^2=\sigma_{sm}^2$ of the broad case through moment matching. To do this, we perform linear interpolation between the variances of the LLN and Softmax Attention by injecting uncorrelated Gaussian inputs to both attentions and measuring their output variances according to the following equation:
\begin{equation}
    \sigma_{sm}^2 = \sigma_{lln}^2 = a \Tilde{\sigma}^2 + b
\end{equation}
Once we have determined the values for $a$ and $b$, we can substitute them into \Cref{eq:lln_alpha_beta} to obtain the optimal values for $\alpha$ and $\beta$.

\begin{figure}
\centering
 \begin{subfigure}[t]{0.49\textwidth}
 \centering
    \includegraphics[width=1.0\columnwidth]{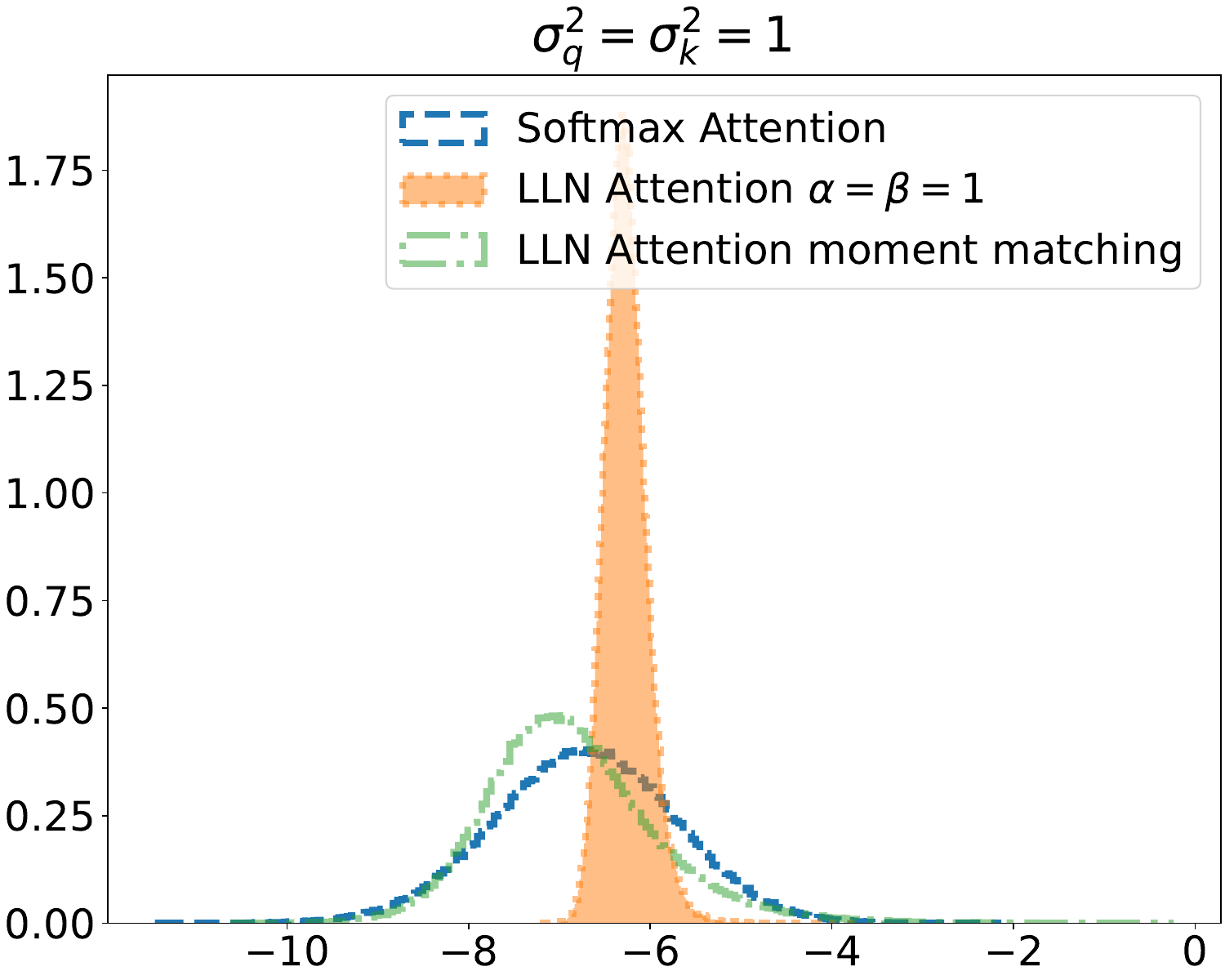}
 \end{subfigure}
 \hfill
 \begin{subfigure}[t]{0.49\textwidth}
 \centering
    \includegraphics[width=1.00\columnwidth]{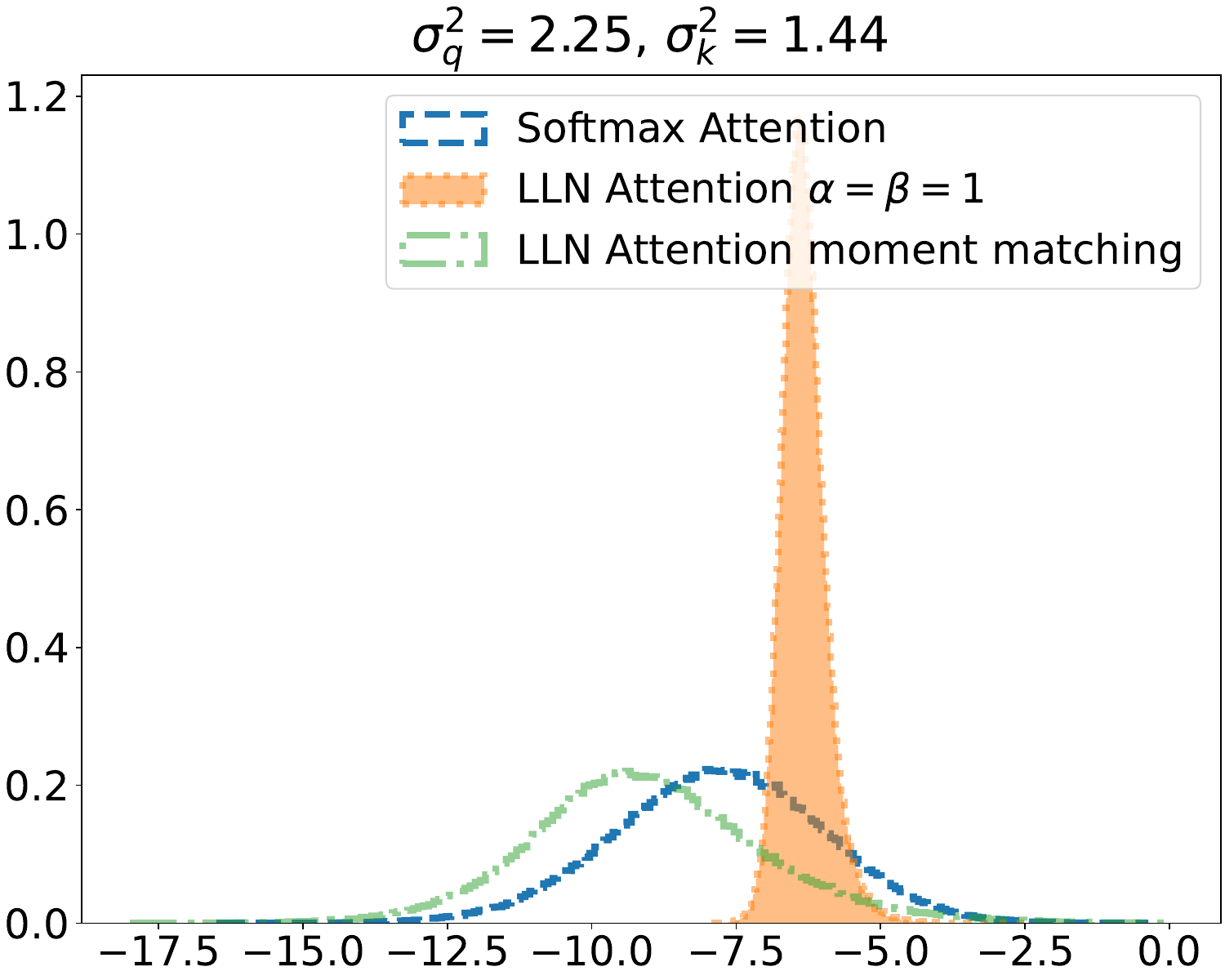}
 \end{subfigure}
 \caption{Histogram of the SA and LLN Attention before and after performing the moment matching.}
  \label{fig:lln_histogram}
\end{figure}

As depicted in \Cref{fig:lln_moment_matching}, the variance of the LLN Attention without moment matching (i.e., $\alpha=\beta=1$) is much smaller than that of the Softmax Attention and exhibits a nearly linear trend. However, the variance of the LLN Attention with moment matching approximates that of the Softmax Attention. Additionally, the histogram shown in \Cref{fig:lln_histogram} suggests that the LLN Attention distribution closely follows that of the Softmax Attention. Despite that, the two distributions have slightly different means because we only match their variances.

\subsection{Experiments}\label{sec.appendinx_experiments}
In this section, we present more experimental results and ablation studies of our method.

\subsubsection{Pre-training of RoBERTa model}\label{sec.appendinx_experiments_roberta}
\begin{figure}
\centering
 \begin{subfigure}[t]{0.49\textwidth}
 \centering
    \includegraphics[width=1.0\columnwidth]{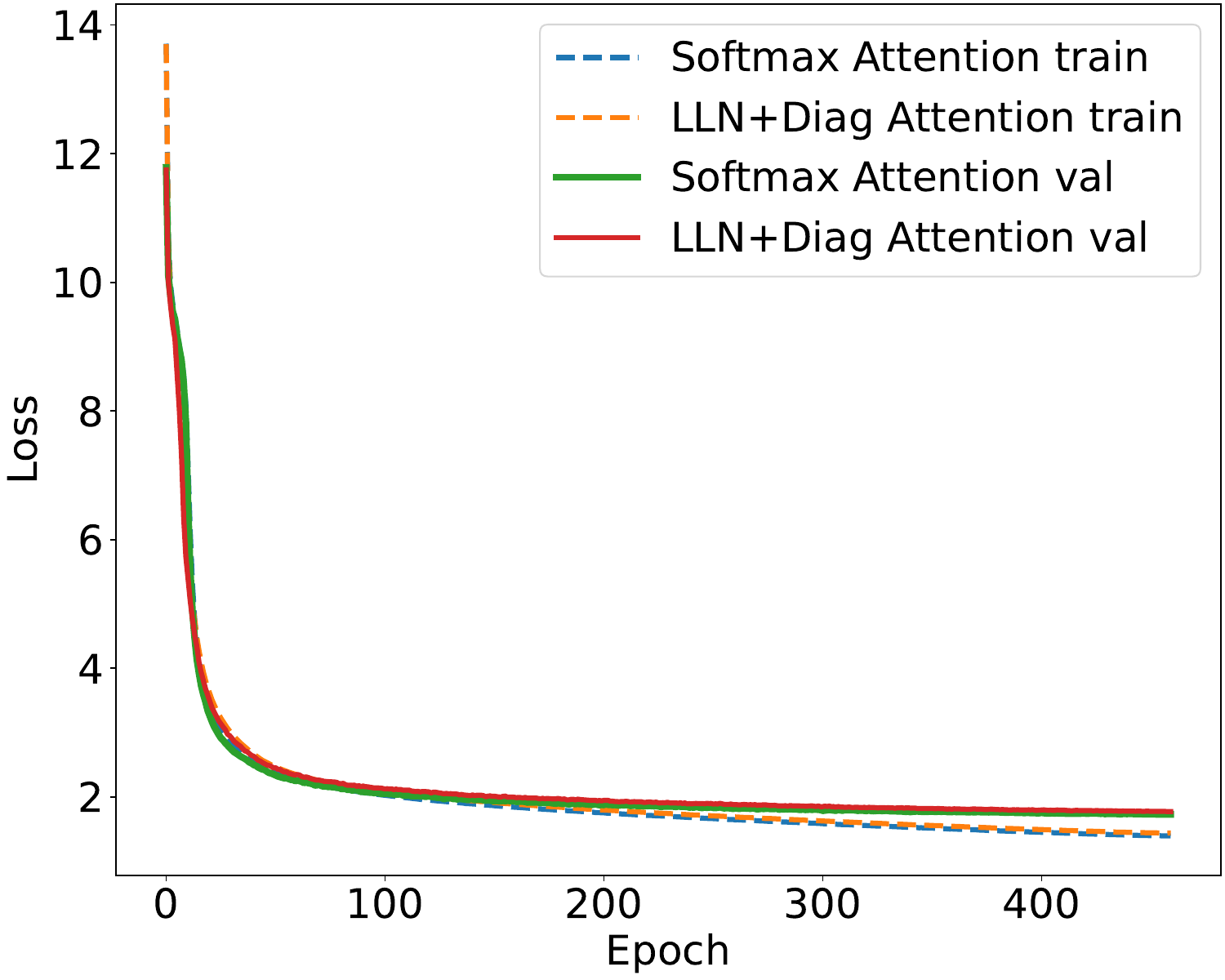}
    \caption{}
\label{fig:roberta_pretraining}
 \end{subfigure}
 \hfill
 \begin{subfigure}[t]{0.49\textwidth}
 \centering
    \includegraphics[width=1.00\columnwidth]{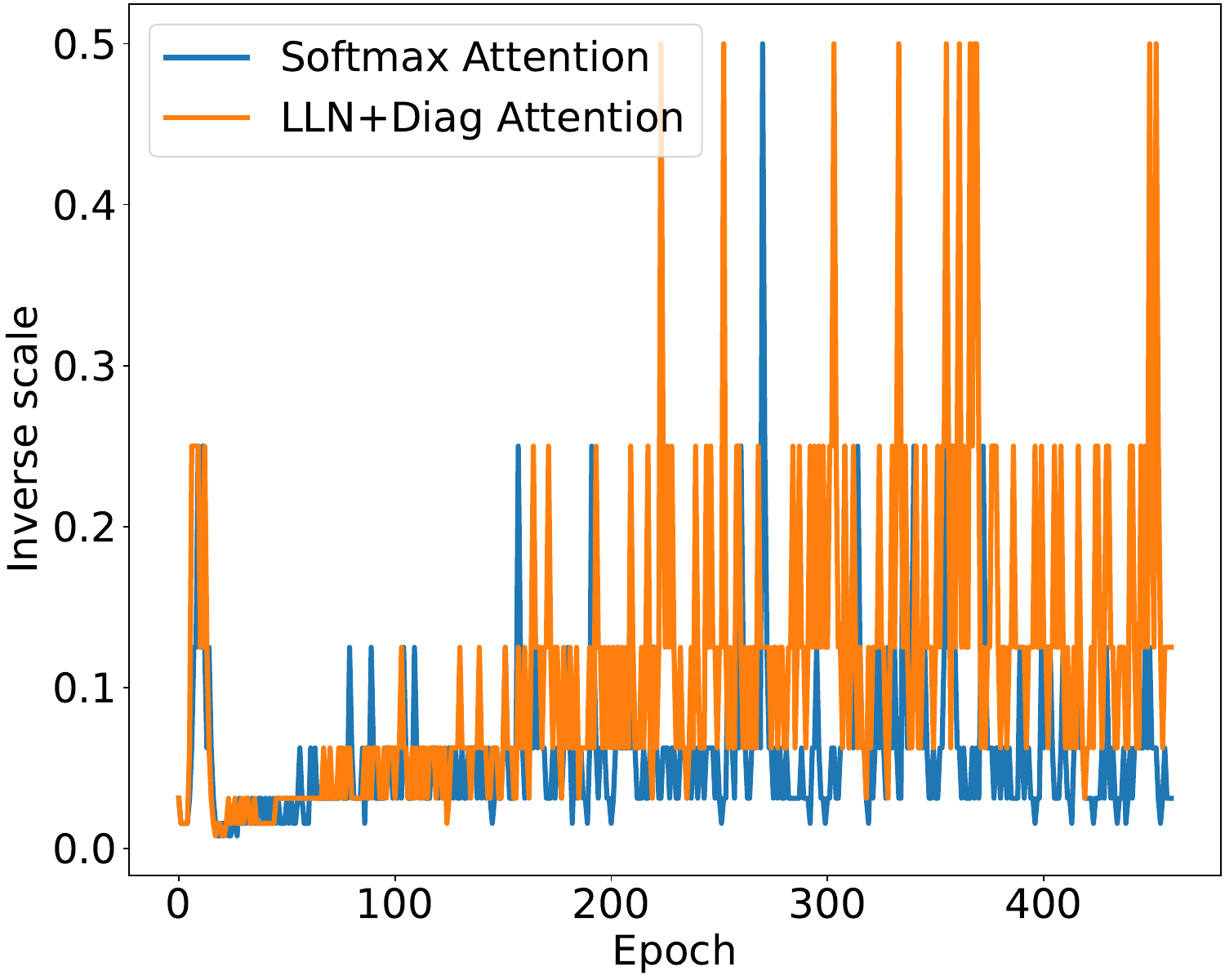}
    \caption{}
 \label{fig:roberta_train_scale}
 \end{subfigure}
 \caption{(a) Training and validation loss comparison of RoBERTa-base model pre-training using LLN Attention and SA. (b) Inverse of the loss scale during training of RoBERTa-base model.}
\end{figure}

We train the bidirectional RoBERTa-base encoder model \citep{Roberta_robustify} using LLN Attention on the WikiText-103 corpus \citep{Wikitext-103}. During pre-training, we monitor the convergence of the model and compare its performance to the SA model. In \Cref{fig:roberta_pretraining} we show the training and validation loss of the RoBERTa-base model during pre-training with LLN Attention, as well as its comparison to SA. The loss curve of LLN Attention closely follows the SA, indicating similar convergence behavior. We used the Fairseq framework \citep{fairseq_facebook} for all experiments with the default configuration and hyperparameters of the RoBERTa-base model.\footnote{https://github.com/facebookresearch/fairseq/blob/main/examples/roberta/README.md}

We perform the training with FP16 precision, which can cause instability during training. To test the stability of the training, we also log the inverse loss scale parameter \Cref{fig:roberta_train_scale}. Spikes in the plot indicate a decrease in the loss scale due to large gradients. As can be seen from the figure, the maximum inverse scale of LLN Attention does not exceed that of the SA, which is important to ensure similar stability during training as with SA.

\subsubsection{Image classification}
To test LLN Attention on Vision task, we evaluate it on the Vision Transformer model using vit-pytorch \footnote{https://github.com/lucidrains/vit-pytorch} code base. Our ViT model consists of twelve layers and 128 embedding sizes. We train this model for 100 epochs on Dogs vs Cats dataset \footnote{https://www.kaggle.com/competitions/dogs-vs-cats-redux-kernels-edition/data} with LLN and Softmax Attention. The results in \Cref{tab:vit_dog_cats} show that LLN Attention performs on par with SA while outperforming the Linformer \citep{linformer_Wang_2020} method.

\begin{table}[ht]
\centering
\begin{tabular}{|c|c|c|}
\hline
Softmax & LLN+Diag & Linformer \\ \hline
81.37   & 81.72    & 79.89     \\ \hline
\end{tabular}
\caption{Accuracy [\%] of the ViT model trained on Dogs vs Cats dataset with Softmax, LLN (ours) and Linformer \citep{linformer_Wang_2020} Attention.}
\label{tab:vit_dog_cats}
\end{table}

\subsubsection{Long Range Arena}

\begin{table}[ht]
\centering
\begin{tabular}{|l|ccccc|ccccc|}
\hline
& \multicolumn{5}{c|}{Time{[}s{]}}                                                                                        & \multicolumn{5}{c|}{Memory{[}Mb{]}}                                                                                   \\ \hline
\hline
method      & \multicolumn{1}{c|}{TC}    & \multicolumn{1}{c|}{LO}   & \multicolumn{1}{c|}{RE}    & \multicolumn{1}{c|}{PF}   & IC    & \multicolumn{1}{c|}{TC}    & \multicolumn{1}{c|}{LO}   & \multicolumn{1}{c|}{RE}   & \multicolumn{1}{c|}{PF}   & IC   \\ \hline
Softmax    & \multicolumn{1}{c|}{21468} & \multicolumn{1}{c|}{5905} & \multicolumn{1}{c|}{21866} & \multicolumn{1}{c|}{6754} & 13228 & \multicolumn{1}{c|}{17108} & \multicolumn{1}{c|}{4458} & \multicolumn{1}{c|}{8934} & \multicolumn{1}{c|}{4817} & 9632 \\ \hline
Reformer   & \multicolumn{1}{c|}{4610}  & \multicolumn{1}{c|}{2439} & \multicolumn{1}{c|}{4714}  & \multicolumn{1}{c|}{4694} & 8737  & \multicolumn{1}{c|}{3261}  & \multicolumn{1}{c|}{1631} & \multicolumn{1}{c|}{3008} & \multicolumn{1}{c|}{3258} & 6514 \\ \hline
Performer  & \multicolumn{1}{c|}{3456}  & \multicolumn{1}{c|}{1966} & \multicolumn{1}{c|}{3761}  & \multicolumn{1}{c|}{3553} & 13169 & \multicolumn{1}{c|}{2176}  & \multicolumn{1}{c|}{1122} & \multicolumn{1}{c|}{2178} & \multicolumn{1}{c|}{2180} & 4353 \\ \hline
Skyformer  & \multicolumn{1}{c|}{4523}  & \multicolumn{1}{c|}{2970} & \multicolumn{1}{c|}{5602}  & \multicolumn{1}{c|}{5240} & 9347  & \multicolumn{1}{c|}{3068}  & \multicolumn{1}{c|}{1697} & \multicolumn{1}{c|}{2974} & \multicolumn{1}{c|}{4041} & 8079 \\ \hline
LLN + Diag & \multicolumn{1}{c|}{3043}  & \multicolumn{1}{c|}{1774} & \multicolumn{1}{c|}{3135}  & \multicolumn{1}{c|}{3042} & 4053  & \multicolumn{1}{c|}{1641}  & \multicolumn{1}{c|}{821}  & \multicolumn{1}{c|}{1586} & \multicolumn{1}{c|}{1639} & 3276 \\ \hline
\end{tabular}
\caption{Comparison of memory[Mb] and running time [s] of LLN Attention with Reformer\citep{reformer}, Performer\citep{performer} and Skyformer\citep{skyformer_Chen_2021} linear attention methods and SA baseline.}
\label{tab:lra_memory}
\end{table}

We use the Long Range Arena (LRA) \citep{lra-google} benchmark to evaluate LLN Attention on longer sequences. LRA benchmark requires a sequence length between 1k and 4k, depending on the task. To that end, we used a code base provided by Skyformer \citep{skyformer_Chen_2021} \footnote{https://github.com/pkuzengqi/Skyformer}. We compare the LRA score in addition to the memory and computation complexity of LLN Attention with Reformer\citep{reformer}, Performer\citep{performer}, and Skyformer\citep{skyformer_Chen_2021} linear methods as well as regular SA. According to the \Cref{tab:lra_memory}, LLN Attention requires much less memory and time compared to other methods while achieving a similar average LRA score as SA \Cref{tab:lra_score}.

\begin{table}[ht]
\centering
\begin{tabular}{|l|c|c|c|c|c|c|}
\hline
method     & Text (4k) & ListOps (2k) & Retrieval (4k) & Pathfinder (1k) & Image (1k) & AVG            \\ \hline \hline
Softmax    & 60.41     & 38.05        & 79.95          & 71.3            & 37.2       & 57.38          \\ \hline
Reformer   & 61.27     & 37.05        & 78.74          & 67.23           & 44.04      & 57.67          \\ \hline
Performer  & 57.85     & 37.8         & 80.5           & 62.58           & 37.56      & 55.26          \\ \hline
Skyformer  & 60.88     & 39.36        & 81.54          & 70.23           & 32.64      & 56.93          \\ \hline
LLN + Diag & 60.72     & 38.91        & 81.21          & 69.81           & 38.65      & \textbf{57.86} \\ \hline
\end{tabular}
\caption{LRA score of LLN Attention with Reformer\citep{reformer}, Performer\citep{performer} and Skyformer\citep{skyformer_Chen_2021} linear attention methods and SA baseline.}
\label{tab:lra_score}
\end{table}

\subsubsection{LLN Attention concentration - ablation study}

\begin{figure}[t]
\centering
 \begin{subfigure}[t]{0.49\textwidth}
 \centering
    \includegraphics[width=1.0\columnwidth]{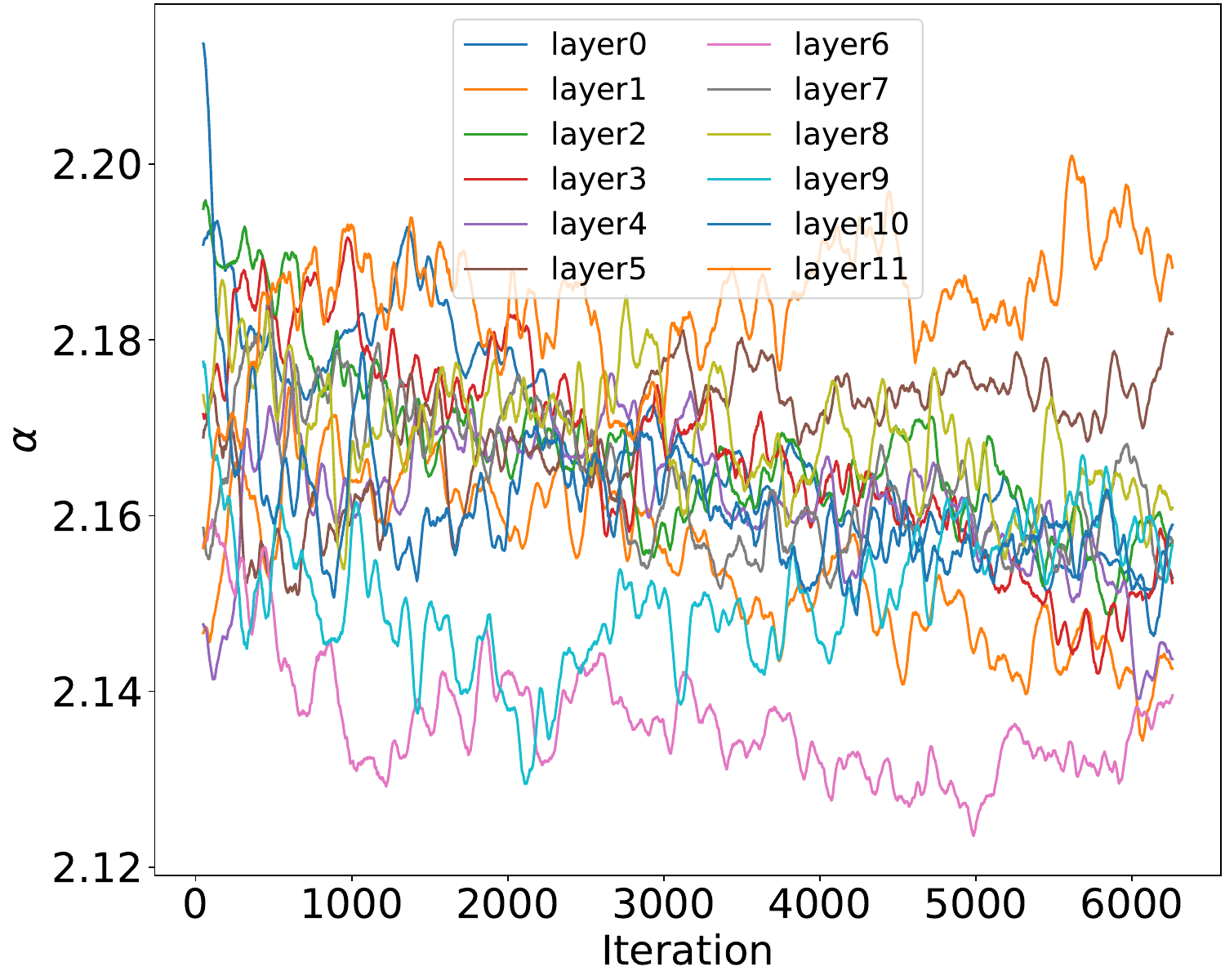}
    \caption{}
\label{fig:alpha_mm}
 \end{subfigure}
 \hfill
 \begin{subfigure}[t]{0.49\textwidth}
 \centering
    \includegraphics[width=1.00\columnwidth]{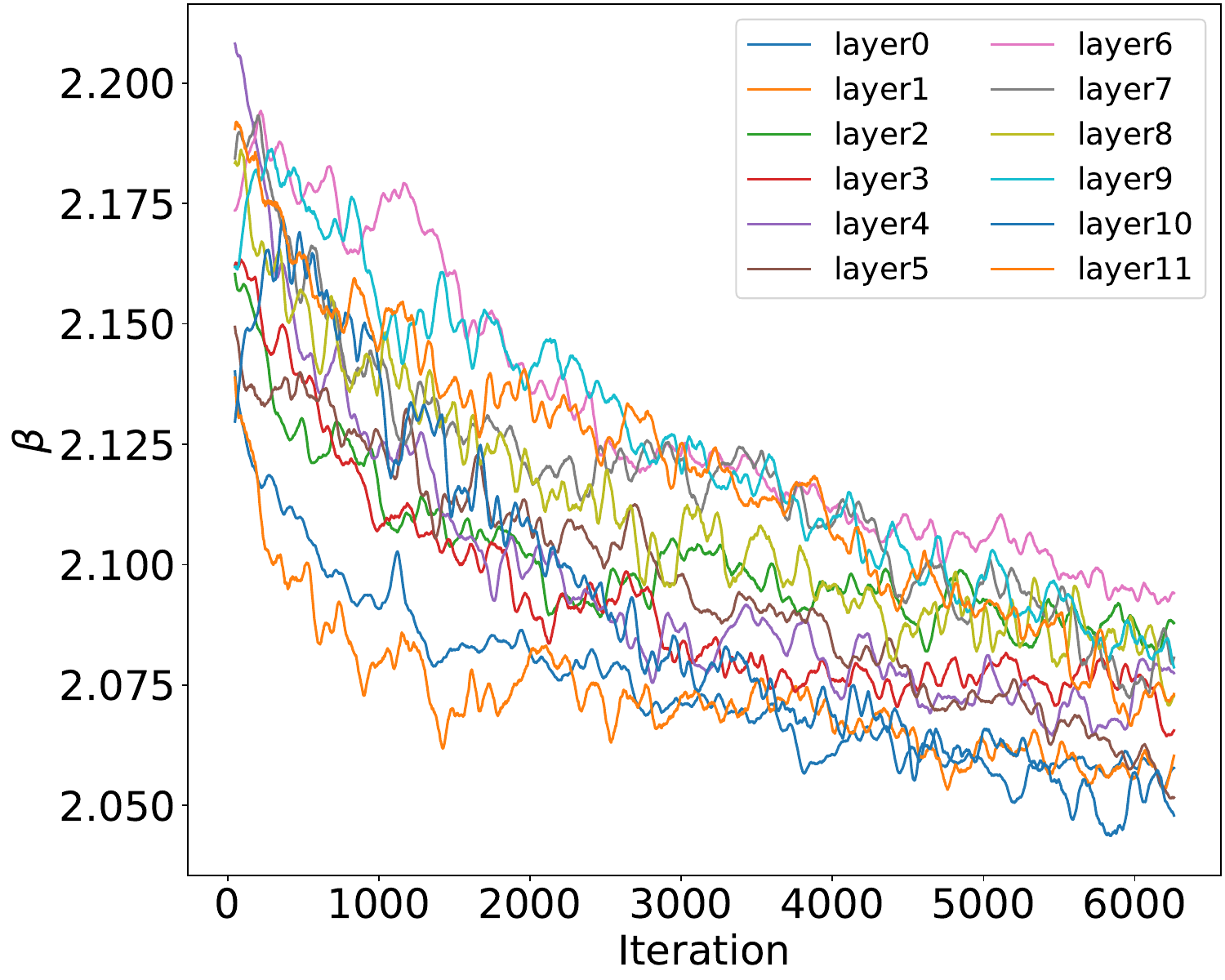}
    \caption{}
 \label{fig:beta_mm}
 \end{subfigure}
 \caption{Change in the $\alpha$ parameter (a) and  $\beta$ (b) during training of ViT model on Dogs vs Cats dataset.}
  \label{fig:mm_alpha_beta}
\end{figure}

\begin{figure}
\centering
 \begin{subfigure}[t]{0.49\textwidth}
 \centering
    \includegraphics[width=1.0\columnwidth]{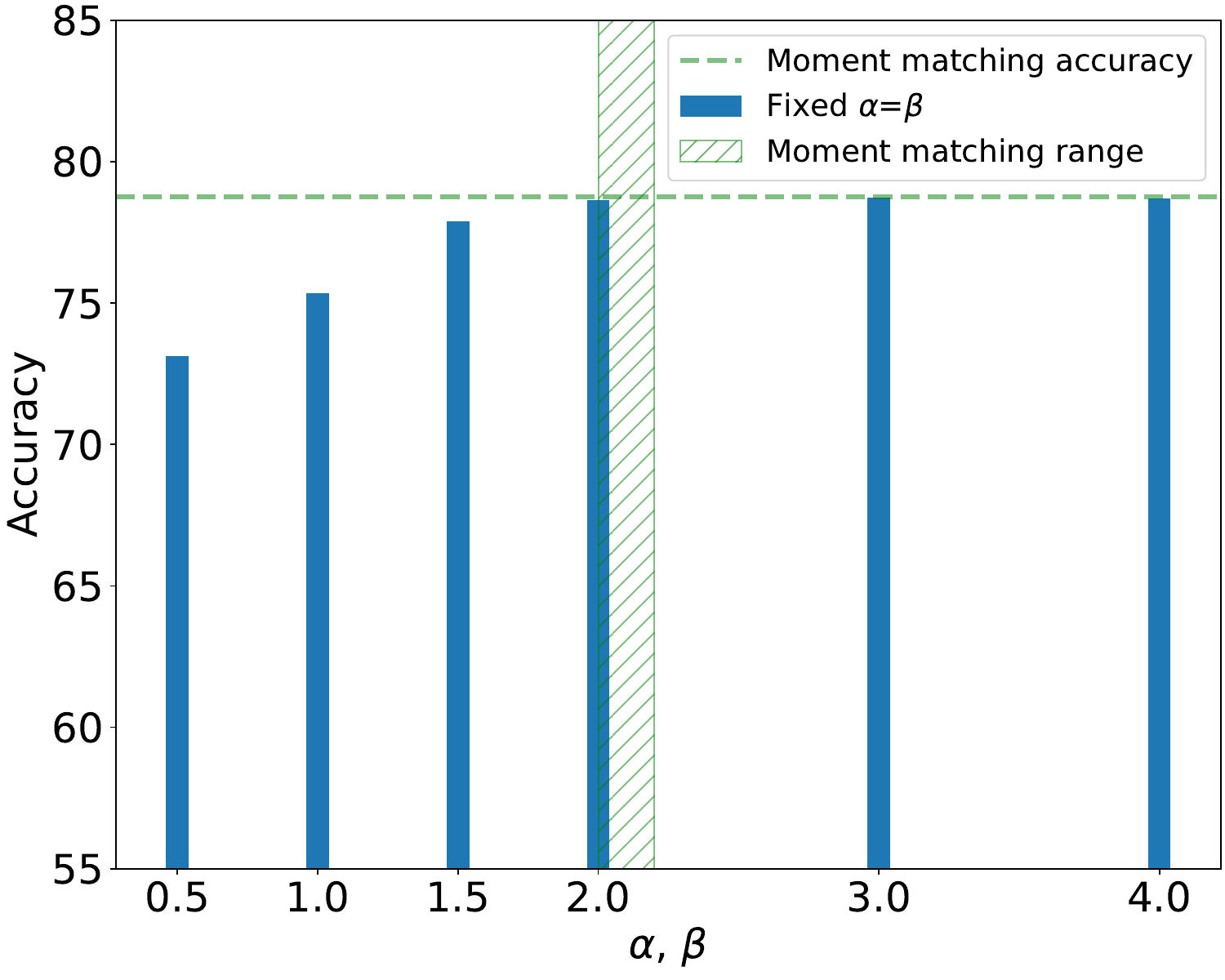}
    \caption{}
\label{fig:alpha_beta_ablation}
 \end{subfigure}
 \hfill
 \begin{subfigure}[t]{0.49\textwidth}
 \centering
    \includegraphics[width=1.00\columnwidth]{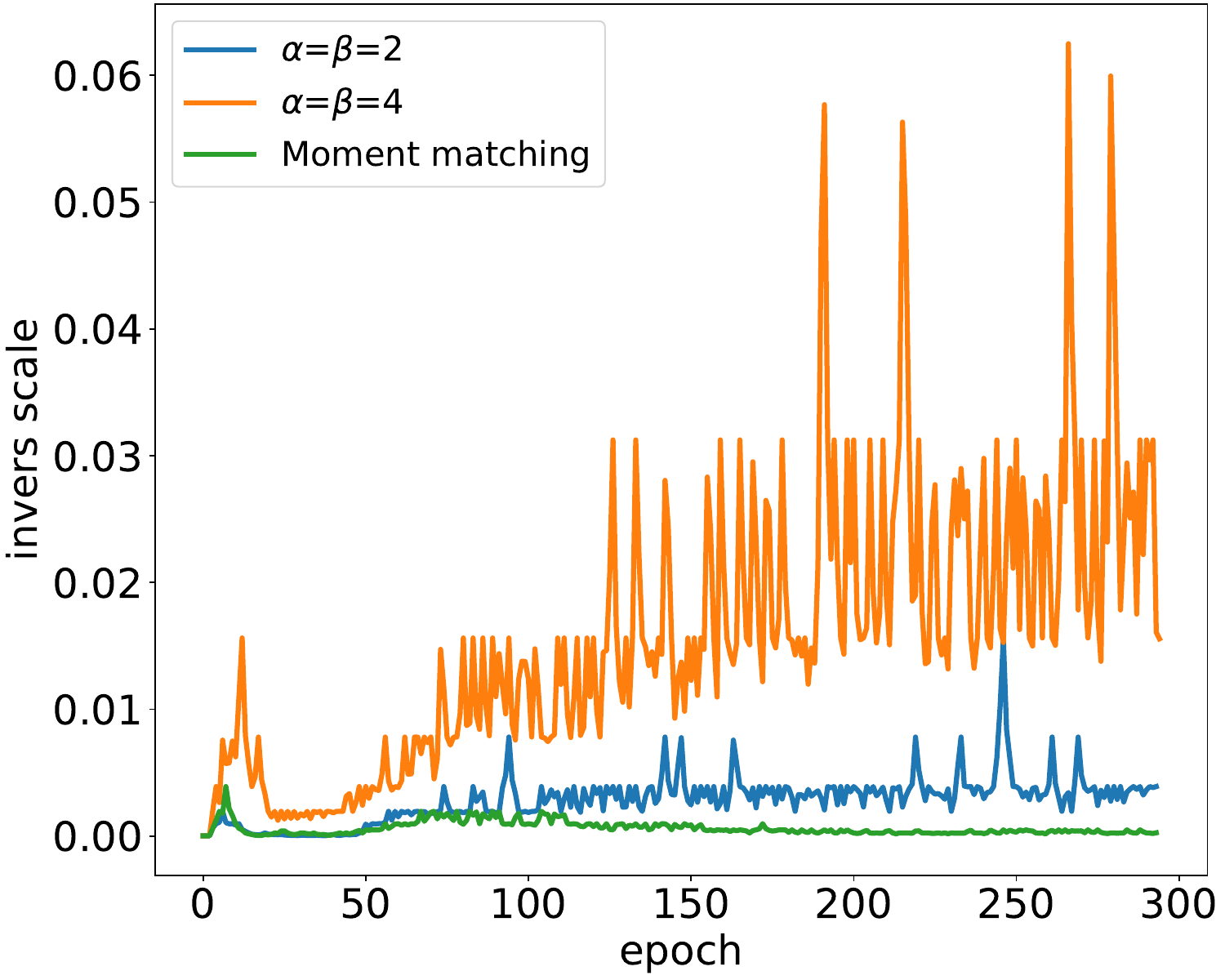}
    \caption{}
 \label{fig:deit_inverse_scale}
 \end{subfigure}
 \caption{(a) Accuracy of ViT trained on Dogs vs Cats dataset with LLN Attention and different values of $\alpha$ and $\beta$. (b) Inverse of the loss scale during training of deit-tiny model for different fixed values of $\alpha$ and $\beta$ as well as for moment matching.}
\end{figure}

In this section, we analyze the impact of the LLN Attention temperature of the Vision Transformer (ViT) model trained on the Dogs vs Cats dataset. First, we record the values of $\alpha$ and $\beta$ produced by the moment matching procedure during training. According to \Cref{fig:mm_alpha_beta}, the values of $\alpha$ and $\beta$ obtained during moment matching lay within the range of $(2; 2.2)$. Furthermore, since the temperature of the LLN Attention, as defined in \Cref{eq:lln_attn_temperature}, decreases as the hyper-parameters $\alpha$ and $\beta$ increase. To assess the influence of the temperature, we train the model with various fixed values of hyper-parameters $\alpha$ and $\beta$ and record the resulting accuracy. In \Cref{fig:alpha_beta_ablation}, we see that when $\alpha$ and $\beta$ values are smaller than the moment matching range, i.e., less than $2$, the LLN Attention concentration is insufficient due to the high temperature, leading to accuracy degradation. Conversely, when $\alpha$ and $\beta$ values lay within the range of moment matching values or larger ($\alpha, \beta \geq 2$), the concentration is sufficient for the model to achieve the desired accuracy.

We highlight the risks associated with surpassing the moment matching range by increasing $\alpha$ and $\beta$. In particular, larger values of these parameters may risk the stability of the training process due to increased gradients, a concern that becomes especially noticeable when training models in Float16 format. The risk of utilizing the Float16 data type stems from the reduced precision, smaller dynamic range, risks of gradient overflow, and the requirement to maintain the loss scaling. Moreover, the lower precision of Float16 may result in information loss during computations and numerical instability.

Accordingly, exceeding the moment matching values of $\alpha$ and $\beta$ is practically undesirable, particularly in the context of training with Float16. In \Cref{fig:deit_inverse_scale}, we illustrate this phenomenon by presenting the loss scale during the training of the deit-tiny model\citep{deit_touvron} in the Float16 format. We see that for large values of $\alpha=\beta=4$, the inverse of the loss scale is significantly larger, compared to $\alpha=\beta=2$, indicating increased gradients and the potential of training instability or even failure. Therefore, to achieve the desired accuracy and allow stable training, it is crucial to maintain the temperature in the "sweet spot" specified by the moment matching values.

\end{document}